%% file: main.tex
\pdfoutput=1

\documentclass[11pt]{article}

\usepackage[]{acl}

\usepackage{times}
\usepackage{latexsym}
\usepackage{array}
\usepackage{enumitem}

\usepackage[T1]{fontenc}
\usepackage[utf8]{inputenc}
\usepackage{microtype}

\input{preamble}

\usepackage{subfigure}
\usepackage{nicefrac}
\usepackage{svg}
\DeclareMathAlphabet{\mathbcal}{OMS}{cmsy}{b}{n}
\usepackage{amsmath}
\usepackage{mathrsfs}
\usepackage{comment}

\usepackage{bm}
\usepackage{bbm}
\usepackage{amssymb}
\usepackage{enumitem}

\usepackage{placeins}

\usepackage{multirow}
\usepackage{adjustbox}
\usepackage{booktabs}

\usepackage{graphbox}
\usepackage{graphicx}
\usepackage{booktabs}

\DeclareMathAlphabet{\mathbcal}{OMS}{cmsy}{b}{n}
\usepackage{amsmath}

\usepackage{comment}
\usepackage{bm}
\usepackage{bbm}
\usepackage{amssymb}
\usepackage{enumitem}

\usepackage{wrapfig,lipsum,booktabs}

\usepackage[strict]{changepage}
\usepackage{float}

\usepackage{framed}

\definecolor{formalshade}{rgb}{0.95,0.95,1}

\newcommand{\lla}{LLaMA-3.1-8B-Instruct }
\newcommand{\gpt}{GPT-4o-mini }
\newcommand{\none}{No-Retrieval}
\newcommand{\rand}{Random-Retrieval}
\newcommand{\user}{LaMP}

\newcommand{\tone}{User Product Review Generation}
\newcommand{\ttwo}{Hotel Experience Generation}
\newcommand{\tthr}{Stylized Feedback Generation}
\newcommand{\tfou}{Multi-lingual Review Generation}

\newcommand{\tfiv}{User Product Review Title Generation}
\newcommand{\tsix}{Hotel Experience Summary Generation}
\newcommand{\tsev}{Stylized Feedback Title Generation}
\newcommand{\teig}{Multi-lingual Review Title Generation}

\newcommand{\tnin}{User Product Review Ratings}
\newcommand{\tten}{Hotel Experience Ratings}
\newcommand{\tele}{Stylized Feedback Ratings}
\newcommand{\ttwe}{Multi-lingual Product Ratings}

\newcommand{\revise}[1]{\textcolor{black}{#1}}

\title{Personalized Graph-Based Retrieval for Large Language Models}

\author{
\textbf{Steven Au\textsuperscript{1}},
\textbf{Cameron J. Dimacali\textsuperscript{1}},
\textbf{Ojasmitha Pedirappagari\textsuperscript{1}},
\\
 \textbf{Namyong Park \textsuperscript{2}},
 \textbf{Franck Dernoncourt\textsuperscript{3}},
 \textbf{Yu Wang\textsuperscript{4}},
\textbf{Nikos Kanakaris\textsuperscript{5}},
\\
 \textbf{Hanieh Deilamsalehy\textsuperscript{3}}
 \textbf{Ryan A. Rossi\textsuperscript{3}},
 \textbf{Nesreen K. Ahmed\textsuperscript{6}}
\\
 \textsuperscript{1}University of California Santa Cruz,
 \textsuperscript{2}Meta AI
 \textsuperscript{3}Adobe Research,
 \\
 \textsuperscript{4}University of Oregon,
 \textsuperscript{5}University of Southern California,
 \textsuperscript{6}Cisco AI Research
\\
}

\begin{document}
\maketitle

\begin{abstract}

\revise{As large language models (LLMs) continue to evolve, their ability to deliver personalized, context-aware responses holds significant promise for enhancing user experiences. However, most existing personalization approaches rely solely on user history, limiting their effectiveness in cold-start and sparse-data scenarios. We introduce Personalized Graph-based Retrieval-Augmented Generation (PGraphRAG), a framework that enhances personalization by leveraging user-centric knowledge graphs. By integrating structured user information into the retrieval process and augmenting prompts with graph-based context, PGraphRAG improves both relevance and generation quality. We also present the Personalized Graph-based Benchmark for Text Generation, designed to evaluate personalized generation in real-world settings where user history is minimal. Experimental results show that PGraphRAG consistently outperforms state-of-the-art methods across diverse tasks, achieving average ROUGE-1 gains of 14.8\% on long-text and 4.6\% on short-text generation—highlighting the unique advantages of graph-based retrieval for personalization.}

\end{abstract}

\input{text/new_intro}

\input{text/new_benchmark}

\input{text/approach}
\input{text/exp-updated}

\section{Conclusion} \label{sec:conc}

\revise{We presented PGraphRAG, a framework that enhances personalized text generation by integrating user-centric knowledge graphs into retrieval-augmented generation. Unlike prior methods that rely solely on user history, PGraphRAG enriches generation with structured user profiles, enabling adaptive personalization even in sparse data settings. Our experiments show that graph-based retrieval significantly improves performance across diverse tasks, outperforming state-of-the-art baselines. Beyond improved metrics, PGraphRAG introduces a scalable design that generalizes user preferences and adapts to new users through structural retrieval. This work lays a foundation for future personalized LLM systems, particularly in applications requiring robustness to data sparsity, cold starts, and context adaptation.
}

\section{Limitations} 
\revise{While PGraphRAG demonstrates strong performance across personalized generation tasks, there are several considerations that present opportunities for future enhancement.}

\revise{\textbf{Scalability considerations.}  
Although personalization approaches can raise scalability concerns, PGraphRAG is designed for efficient large-scale deployment. It constructs a unified, sparse user-item bipartite graph offline --- i.e., graph construction is a one-time cost, similar to those used in scalable recommender systems. As shown in Table~\ref{tab:graph-stats}, the graph is inherently sparse, enabling efficient storage and indexing. At inference time, rather than retrieving over the entire corpus as in traditional RAG settings, PGraphRAG scopes retrieval to a localized subgraph centered on the input user. This subgraph includes both the user's own interactions and those of neighboring users who share items. Standard retrievers (e.g., BM25 or Contriever) are then applied over this constrained set, significantly reducing search overhead while retaining personalized context. This design keeps runtime and memory usage low and supports scalable deployment across large user bases. In future work, we plan to explore compression techniques and real-time profile updates to further enhance scalability in dynamic environments.}

\revise{\textbf{Graph completeness and data sparsity.}  
While the quality of retrieval can be influenced by the completeness of the user-centric graph, PGraphRAG is explicitly designed to operate under sparse and noisy conditions. Our benchmark includes users with minimal interaction history, yet results show strong performance across tasks compared to baseline methods. This robustness arises from PGraphRAG’s graph-based retrieval strategy, which leverages neighboring nodes to provide relevant contextual signals even when direct user data is limited. Nonetheless, integrating implicit signals (e.g., click rate or engagement time) and developing more resilient retrieval methods for incomplete graphs remains a promising direction for future work.}

\revise{\textbf{Generalization vs. user adaptation.}  
A core challenge lies in developing training strategies that balance individual personalization with generalization across user populations. While our approach augments prompts with structured context, future work may explore personalized fine-tuning or adapter layers to enhance this tradeoff further.}

\revise{\textbf{Static user profiles.}  
Currently, user profiles are treated as static during evaluation. In real-world scenarios, preferences evolve over time. Extending the framework to model temporal dynamics and support profile updates is a promising direction for improving long-term personalization.}

\appendix
\input{text/appendix-up}

\newpage
\bibliography{main}
\bibliographystyle{acl_natbib}

\end{document}

%% file: preamble.tex
\definecolor{thedarkblue}{RGB}{0,0,120} 
\definecolor{mydarkblue}{rgb}{0,0.08,0.45} 
\definecolor{darkblue}{rgb}{0,0.08,180}
\colorlet{TufteRed}{red!80!black}
\definecolor{theblue}{RGB}{0,0,180}
\colorlet{thered}{TufteRed}
\usepackage{microtype}
\usepackage{balance}
\usepackage{tcolorbox}

\usepackage{booktabs}
\usepackage{tabularx}

\usepackage{amsmath,amssymb,amsthm}

\newcommand{\eat}[1]{\ignorespaces}
\usepackage{comment}

\newcommand{\journal}[1]{} 

\usepackage{svg}
\usepackage{tikz}
\usepackage{verbatim}
\usetikzlibrary{arrows}
\usetikzlibrary{shapes,snakes}
\usetikzlibrary{decorations.pathmorphing} 
\usetikzlibrary{fit}					
\usetikzlibrary{backgrounds}	

\usepackage{ragged2e}
\usepackage{multirow}
\usepackage{microtype}
\usepackage{balance}
\usepackage{setspace}

\graphicspath{{./}{./graphics/}}
\newcolumntype{H}{>{\setbox0=\hbox\bgroup}c<{\egroup}@{}}

\newcolumntype{R}[1]{>{\RaggedLeft\arraybackslash}} 
\newcolumntype{L}[1]{>{\RaggedRight\arraybackslash}} 

\AtBeginEnvironment{pmatrix}{\setlength{\arraycolsep}{2pt}}

\DeclareMathOperator{\hugeE}{\mbox{\huge\raise-0.3ex\hbox{E}}}
\DeclareMathOperator{\p}{\mathbb{P}}
\DeclareMathOperator{\hugep}{\mbox{\huge\raise-0.3ex\hbox{$\p$}}}

\usepackage{pgfplotstable}  
\usepackage{booktabs}       
\usepackage{graphicx}       
\usepackage{caption}        
\usepackage{multirow}       
\usepackage{array}          
\usepackage{makecell}       

%% file: text/new_intro.tex
\section{Introduction}

\begin{figure}[t!]
    \centering
    \includegraphics[width=1.0\columnwidth]{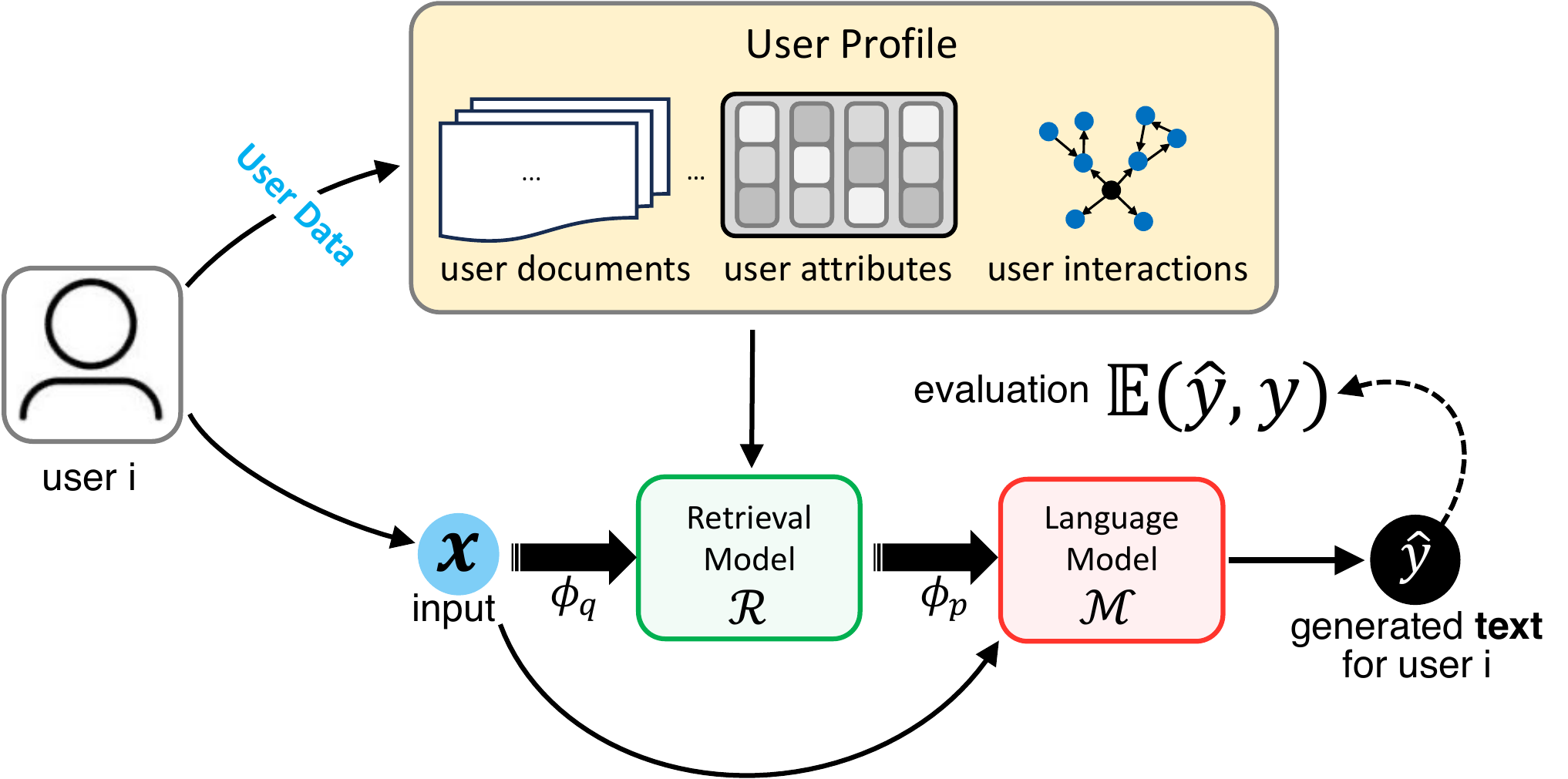}
    \caption{%
    Overview of the proposed PGraphRAG framework. We construct user-centric graphs from user profile and interaction data, then retrieve structured, user-relevant information from the graph. This context is used to condition the language model's generation, producing personalized outputs for user $i$.}
    \label{fig:overview-personalized-graph-based-LLM}
    \vspace{-3mm}
\end{figure}

The rapid advancement of large language models (LLMs) has enabled a wide range of NLP applications, including conversational agents, content generation, and code synthesis. Models like GPT-4~\cite{openai2024gpt4ocard} now power virtual assistants capable of answering complex queries and engaging in multi-turn dialogue~\cite{llm-fewshot}. As these models continue to evolve, their ability to generate personalized, context-aware responses offers new opportunities to enhance user experiences~\cite{salemi2024lamp, huang-personalization}. Personalization enables LLMs to adapt outputs to individual preferences and goals, resulting in richer, more relevant interactions~\cite{zhang2024personalizationlargelanguagemodels}. While personalization has been studied in areas such as information retrieval and recommender systems~\cite{xue-personalizedsearch, DLRM19}, its integration into LLMs for generation tasks remains relatively underexplored.

One of the key challenges in advancing personalized LLMs is the lack of benchmarks that adequately capture the complexities of personalization tasks. Popular natural language processing (NLP) benchmarks (e.g., \cite{wang2018glue}, \cite{10.5555/3454287.3454581}, \cite{gehrmann-etal-2021-gem}) primarily focus on general language understanding and generation, with limited emphasis on personalization. As a result, researchers and practitioners lack standardized datasets and evaluation metrics for developing and assessing models designed for personalized text generation. Recently, efforts such as LaMP~\cite{salemi2024lamp} and LongLaMP~\cite{kumar2024longlampbenchmarkpersonalizedlongform} have begun addressing this gap. LaMP evaluates personalization for tasks like email subject and news headline generation, while LongLaMP extends this to long-text tasks such as email and abstract generation. However, both benchmarks rely exclusively on user history to model personalization. Here, user history typically refers to a set of previously written texts by the same user—such as past reviews, messages, or profile-specific documents—which are used as context to condition the generation.

\begin{figure}[ht]
    \centering
    \includegraphics[width=0.8\columnwidth]{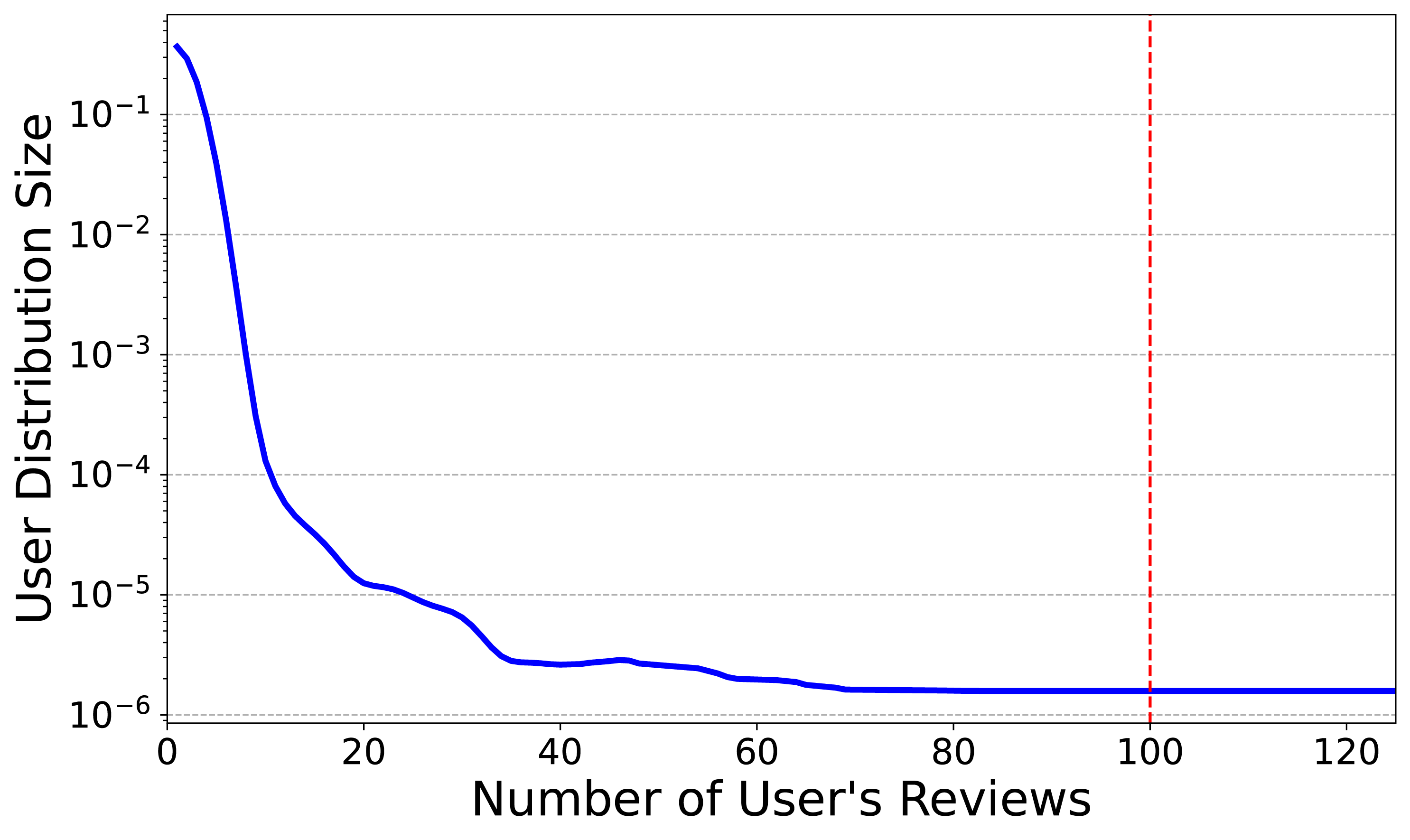} 
    \caption{%
    Distribution of user profile sizes in the Amazon user-product dataset. The vast majority of users have only a few reviews, highlighting the prevalence of sparse profiles. The red vertical line indicates the minimum profile size threshold used in prior benchmarks such as LaMP and LongLaMP.}

    \label{fig:profile_distribution}
    \vspace{-3mm}
\end{figure}

\noindent \textbf{Challenges with Cold-Start Users.} \revise{While leveraging user history is valuable for capturing individual style and preferences, it presents a cold-start challenge: many users have little or no prior data. In fact, as shown in Figure~\ref{fig:profile_distribution}, over 99.99\% of users in the Amazon Reviews dataset have fewer than three interactions. Benchmarks like LaMP and LongLaMP filter out these users by imposing a minimum user profile size threshold to ensure sufficient data for personalization. As a result, they exclude the vast majority of users, making their evaluations less representative of real-world deployment. This design choice leads to model failures when prompts lack sufficient context, often resulting in generic outputs. }

\noindent \textbf{Proposed Approach.} To address these challenges, we propose \emph{Personalized Graph-based Retrieval-Augmented Generation} (PGraphRAG), a novel framework that enhances personalized text generation by leveraging user-centric knowledge graphs. These structured graphs represent user information --- such as interests, preferences, and prior interactions --- in an interconnected graph structure. During inference, PGraphRAG retrieves semantically relevant context from both the user's own profile and neighboring profiles extracted from the graph, and augments the prompt with this information to guide generation. This graph-based approach enables the model to produce contextually appropriate and personalized outputs, even when user history is sparse or unavailable (see Figure~\ref{fig:overview-personalized-graph-based-LLM}).

\revise{Formally, the target task of PGraphRAG is personalized text generation conditioned on user-specific context retrieved from a structured knowledge graph. Given a user query (e.g., a product title or review prompt), the system retrieves relevant entries from the graph-based profile and generates an output tailored to the user's preferences. This setup generalizes personalization beyond pure user text history, enabling context-rich generation even in sparse or cold-start settings.}

\noindent \textbf{Proposed Benchmark.} To evaluate our approach, we introduce the \emph{Personalized Graph-based Benchmark for Text Generation}, a novel evaluation benchmark designed to fine-tune and assess LLMs on twelve personalized text generation tasks, including long- and short-form generation as well as classification. This benchmark addresses the limitations of existing personalized LLM benchmarks by providing datasets that specifically target personalization capabilities in real-world settings where user history is sparse. In addition, it enables a more comprehensive assessment of a model's ability to personalize outputs based on structured user information.

\revise{Our benchmark supports evaluation in sparse-profile settings, and PGraphRAG is designed to retrieve semantically relevant context not only from the user's own profile but also from neighboring profiles extracted from the graph --- enabling effective personalization even when the user has only a single input (e.g., one review in their profile). Empirically, PGraphRAG significantly outperforms LaMP in these low-profile scenarios, demonstrating the advantages of graph-based reasoning over strict reliance on user history.}

Our contributions are summarized as follows:
\begin{enumerate}[topsep=0pt,itemsep=1pt,partopsep=0pt,parsep=0pt]
    \item \textbf{Benchmark.} We introduce the \emph{Personalized Graph-based Benchmark for Text Generation}, consisting of 12 tasks spanning long-form generation, summarization, and classification. To support further research, we release the benchmark publicly.~\footnote{\url{https://github.com/PGraphRAG-benchmark/PGR-LLM}}  
    
    \item \textbf{Method.} We propose \emph{PGraphRAG}, a retrieval-augmented generation framework that addresses the cold-start problem by augmenting generation with structured, user-specific information from a knowledge graph.
    
    \item \textbf{Effectiveness.} We show that PGraphRAG achieves state-of-the-art performance across all tasks in our benchmark, demonstrating the value of graph-based reasoning for personalized text generation.
\end{enumerate}

%% file: text/new_benchmark.tex
\section{Personalized Graph-based Benchmark for LLMs}
\label{sec:benchmark}

\begin{figure}[t]
  \centering
  \includegraphics[width=1.\columnwidth]{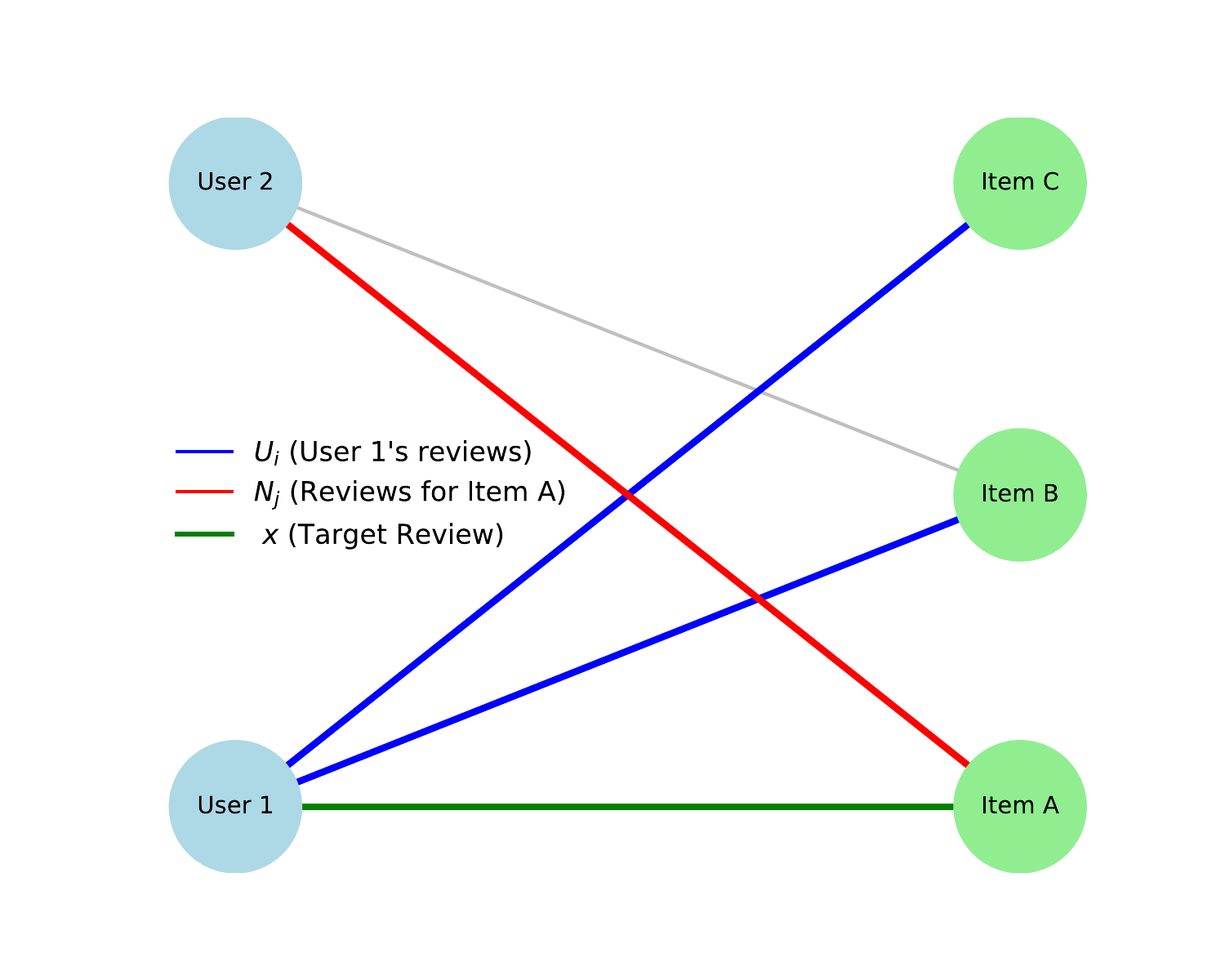}
  \caption{Example of a bipartite user-centric graph $G = (U, V, E)$ showing users, items, and interaction edges (e.g., reviews).}
  \label{fig:bipartite-graph}
\end{figure}

We introduce the \emph{Personalized Graph-Based Benchmark} to evaluate LLMs on their ability to generate personalized outputs across twelve tasks, spanning long-form generation, short-form generation, and ordinal classification. The benchmark is constructed from real-world datasets across multiple domains.

\subsection{Personalized Text Generation: Problem Definition}

Each benchmark instance includes: (1) an input sequence $x$ to the LLM, (2) a target output $y$ the model is expected to generate, and (3) a user profile $P_i$ derived from a structured user-centric graph. Given an input-output pair $(x, y)$ associated with user $i$, the goal is to generate a personalized output $\hat{y}$ that aligns with the semantics and style of $y$, conditioned on the user profile $P_i$.

We assume user context is represented using a bipartite user-centric graph that captures user-item interactions (see Figure~\ref{fig:bipartite-graph} for an illustration). The profile $P_i$ is constructed from this graph and includes both interactions authored by the user and related signals from similar items or neighboring users. The full construction of $P_i$ is detailed in Section~\ref{sec:approach}.

Formally, the personalized generation task is defined as:
\begin{equation}
\hat{y} = \arg\max_{y'} \Pr(y' \mid x, P_i)
\end{equation}
where $x$ is the input query, $y$ is the target output, and $P_i$ denotes the profile of user $i$ derived from a user-item interaction graph. The model generates an output $\hat{y}$ that maximizes the likelihood of personalized text conditioned on the input and user profile. This formulation enables generalization beyond user history by leveraging structured, graph-derived context.

In practice, our framework retrieves a personalized context $\mathcal{R}(P_i) \subseteq P_i$ from the graph to condition generation, yielding the operational objective:
\begin{equation}
\hat{y} = \arg\max_{y'} \Pr(y' \mid x, \mathcal{R}(P_i))
\end{equation}
where $\mathcal{R}(P_i)$ represents the retrieved subset of user- and item-level interactions used as context during generation.

Finally, statistics for all benchmark tasks and their associated graphs are summarized in Table~\ref{tab:task-stats} and Table~\ref{tab:graph-stats}. Additional dataset split details are provided in the appendix.

\input{tables/task_stat}

\input{tables/graph_stats}

\subsection{Task Definitions}

\paragraph{Task 1: \tone.}  
Personalized review text generation has progressed from incorporating user-specific context to utilizing LLMs for producing fluent and contextually relevant reviews and titles~\cite{ni-mcauley-2018-personalized}. This task aims to generate a product review \( i_{\text{text}} \) for a target user, conditioned on their own review title \( i_{\text{title}} \) and a set of additional reviews \( P_i \) from their user profile. We construct this dataset from the Amazon Reviews 2023 corpus~\cite{hou2024bridging}, spanning multiple product categories and used to define a bipartite user-item graph.

\paragraph{Task 2: \ttwo.}  
Hotel reviews often contain rich narratives reflecting personal experiences, making personalization essential to capturing individual preferences and expectations~\cite{kanouchi-etal-2020-may}. This task focuses on generating a personalized hotel experience story \( i_{\text{text}} \), using the target user’s review summary \( i_{\text{title}} \) and contextual reviews \( P_i \). We use the Hotel Reviews dataset, a subset of Datafiniti’s Business Database~\cite{datafiniti2017hotel}, to construct a user-hotel bipartite graph.

\paragraph{Task 3: \tthr.}  
Writing style — influenced by grammar, punctuation, and expression — is deeply personal and often shaped by geographic and cultural factors~\cite{alhafni-etal-2024-personalized}. This task involves generating personalized product feedback \( i_{\text{text}} \), based on the user’s feedback title \( i_{\text{title}} \) and additional feedback samples \( P_i \) from their profile. We utilize the Grammar and Online Product dataset, a subset of the Datafiniti Business corpus~\cite{datafiniti2018grammar}, which reflects stylistic variation across multiple platforms and domains.

\paragraph{Task 4: \tfou.}  
Personalization in multilingual review generation presents unique challenges due to differences in linguistic structures, cultural norms, and stylistic conventions~\cite{cortes-etal-2024-llms}. This task focuses on generating product reviews \( i_{\text{text}} \) in Brazilian Portuguese, using the target user’s review title \( i_{\text{title}} \) and additional reviews \( P_i \) from their profile. We construct this dataset using B2W-Reviews~\cite{real2019b2w}, sourced from Brazil’s largest e-commerce platform.

\paragraph{Task 5: \tfiv.}  
Short text generation for personalized review titles is particularly challenging, requiring the model to summarize sentiment and reflect user-specific phrasing preferences. This task generates a review title \( i_{\text{title}} \) for a given user, using their review text \( i_{\text{text}} \) and additional profile reviews \( P_i \), without relying on parametric user embeddings~\cite{xu-etal-2023-pre-trained}. The dataset is derived from Amazon Reviews~\cite{hou2024bridging}.

\paragraph{Task 6: \tsix.}  
Helping users write summaries of hotel experiences requires distilling detailed narratives into concise summaries that reflect individual preferences~\cite{kamath-etal-2024-generating-hotel}. This task generates a hotel experience summary \( i_{\text{title}} \) based on the user’s full experience text \( i_{\text{text}} \) and additional hotel reviews \( P_i \). We use the Hotel Reviews dataset from the Datafiniti Business Database~\cite{datafiniti2017hotel}.

\paragraph{Task 7: \tsev.}  
Stylized feedback summarization aims to capture individual voice and tone in generating short-form feedback. This task benchmarks stylized opinion generation across domains such as music, groceries, and household items~\cite{iso-etal-2024-noisy-pairing}. The model generates the target user’s feedback title \( i_{\text{title}} \) based on their full feedback text \( i_{\text{text}} \) and additional feedback \( P_i \) from similar users. The dataset is built from the Datafiniti Products dataset~\cite{datafiniti2018grammar}.

\paragraph{Task 8: \teig.}  
Multilingual short-text personalization adds further complexity, particularly in Brazilian Portuguese, where style and syntax vary significantly across users~\cite{scalercio-etal-2024-enhancing}. This task generates a personalized review title \( i_{\text{title}} \) using the user’s full review text \( i_{\text{text}} \) and contextual examples \( P_i \) from their graph neighborhood. Data: B2W-Reviews~\cite{real2019b2w}.

\paragraph{Task 9: \tnin.}  
Predicting personalized product ratings involves understanding sentiment, user bias, and historical feedback. This task formulates rating prediction as an ordinal classification problem, where the model predicts \( i_{\text{rating}} \in \{1, 2, 3, 4, 5\} \) based on the user’s review text \( i_{\text{text}} \), title \( i_{\text{title}} \), and additional profile context \( P_i \). The dataset is constructed from Amazon Reviews~\cite{hou2024bridging}.

\paragraph{Task 10: \tten.}  
Hotel ratings often reflect nuanced factors such as location, cleanliness, and service. This task models hotel experience rating \( i_{\text{rating}} \) prediction as a classification problem based on the user’s review story \( i_{\text{text}} \), summary \( i_{\text{title}} \), and surrounding review context \( P_i \). Data: Datafiniti Hotel Reviews~\cite{datafiniti2017hotel}.

\paragraph{Task 11: \tele.}  
Cross-domain sentiment prediction explores how writing quality and sentiment expression vary across platforms~\cite{yu-etal-2021-cross}. This task assigns a numerical feedback rating \( i_{\text{rating}} \) to a stylized user review using the input review text \( i_{\text{text}} \), review title \( i_{\text{title}} \), and personalized context \( P_i \). The dataset is taken from the Datafiniti Product Database on Grammar and Online Product Reviews~\cite{datafiniti2018grammar}.

\paragraph{Task 12: \ttwe.}  
While sentence-level sentiment classification in Portuguese has seen success~\cite{de-araujo-etal-2024-chatgpt}, this task extends to full review-level sentiment modeling in a multilingual setting. The model predicts a Portuguese user-product rating \( i_{\text{rating}} \) using both the review text \( i_{\text{text}} \), the title \( i_{\text{title}} \), and additional user-item interactions \( P_i \). We construct this dataset using B2W-Reviews~\cite{real2019b2w}.

%% file: tables/task_stat.tex
\begin{table*}[h!]
    \centering
    \begin{adjustbox}{width=1.0\textwidth}
        \begin{tabular}{lccccc} 
        \toprule
            \textbf{Task} & \textbf{Type} & \textbf{Avg. Input Length} & \textbf{Avg. Output Length} & \textbf{Avg. Profile Size} & \textbf{\# Classes} \\
        \midrule
            User-Product Review Generation & Long Text Generation & $3.754\pm2.71$ & $47.90\pm19.28$ & $1.05\pm0.31$ & - \\
            Hotel Experiences Generation & Long Text Generation & $4.29\pm2.57$ & $76.26\pm22.39$ & $1.14\pm0.61$ & - \\
            Stylized Feedback Generation & Long Text Generation & $3.35\pm2.02$ & $51.80\pm20.07$ & $1.09\pm0.47$ & - \\
            Multilingual Product Review Generation & Long Text Generation & $2.9\pm2.40$ & $34.52\pm12.55$ & $1.08\pm0.33$ & - \\
        \midrule
            User-Product Review Title Generation & Short Text Generation & $30.34\pm37.95$ & $7.02\pm1.14$ & $1.05\pm0.31$ & - \\
            Hotel Experiences Summary Generation & Short Text Generation & $90.40\pm99.17$ & $7.64\pm0.92$ & $1.14\pm0.61$ & - \\
            Stylized Feedback Title Generation & Short Text Generation & $37.42\pm38.17$ & $7.16\pm1.11$ & $1.09\pm0.47$ & - \\
            Multilingual Product Review Title Generation & Short Text Generation & $22.17\pm20.15$ & $7.15\pm1.09$ & $1.08\pm0.33$  & - \\
        \midrule
            User-Product Review Ratings & Ordinal Classification & $34.10\pm38.66$  & - & $1.05\pm0.31$ & 5 \\
            Hotel Experiences Ratings & Ordinal Classification & $94.69\pm99.62$  & - & $1.14\pm0.61$ & 5 \\
                Stylized Feedback Ratings & Ordinal Classification & $40.77\pm38.69$ & - & $1.09\pm0.47$ & 5 \\
            Multilingual Product Ratings & Ordinal Classification & $25.15\pm20.75$ & - & $1.08\pm0.33$  & 5 \\
        \bottomrule
        \end{tabular}
    \end{adjustbox}
    \caption{
    Data statistics for the PGraphRAG Benchmark across the four datasets. For each task, we report the average input and output lengths (in words), measured on the test set using BM25-based retrieval with GPT. The average profile size indicates the number of reviews per user used for personalization.
    }
    \label{tab:task-stats}
\end{table*}

%% file: tables/graph_stats.tex
\begin{table*}[h!]
    \centering
    \large
    \begin{adjustbox}{width=0.8\textwidth}
        \begin{tabular}{lcccc} 
        \toprule
        \textbf{Dataset} & \textbf{Users} & \textbf{Items} & \textbf{Edges/Reviews} & \textbf{Average Degree} \\
        \midrule
        User-Product Review Graph & 184,771 & 51,376 & 198,668 & 1.68 \\
        Hotel Experiences Graph & 15,587 & 2,975 & 19,698 & 2.12 \\
        Stylized Feedback Graph & 58,087 & 600 & 71,041 & 2.42 \\
        Multilingual Product Review Graph & 112,993 & 55,930 & 131,075 & 1.55 \\
        \bottomrule
        \end{tabular}
    \end{adjustbox}
    \caption{
    Graph statistics for the datasets used in the personalized tasks. Each row reports the number of users, items, and edges (i.e., reviews), as well as the average degree of the resulting user-centric bipartite graph. The four graphs correspond to: User-Product, Multilingual Product, Stylized Feedback, and Hotel Experiences.
    }
    \label{tab:graph-stats}
\end{table*}

%% file: text/approach.tex
\section{The PGraphRAG Framework} \label{sec:approach}

Personalizing LLMs in real-world settings requires addressing two key challenges: (1) user profiles are often sparse or unavailable, and (2) incorporating additional user-related context must remain relevant, efficient, and scalable. To tackle these issues, PGraphRAG leverages structured user-centric knowledge graphs for context construction, and combines this with retrieval-augmented prompting. This design enables the model to generalize beyond parametric user embeddings or history-based filtering by dynamically retrieving relevant signals from graph-based user profiles that extend beyond the user’s direct history.

Here, we present \emph{PGraphRAG}, our proposed framework for personalizing large language models (LLMs) through graph-based retrieval augmentation. PGraphRAG enhances generation by conditioning a shared LLM on structured, user-specific context extracted from a user-centric knowledge graph. This enables tailored and context-aware outputs, especially in sparse or cold-start scenarios.

PGraphRAG leverages a bipartite user-centric graph $G = (U, V, E)$ to incorporate contextual signals beyond direct user history. We represent user context as a bipartite graph, where $U$ is the set of user nodes, $V$ the set of item nodes, and $E$ the set of interaction edges (see Figure~\ref{fig:bipartite-graph} for an illustration). An edge $(i, j) \in E$ corresponds to an interaction between user $i$ and item $j$, such as a review that includes metadata like text, title, and rating. The user profile $P_i$ consists of the set of reviews written by user $i$, along with reviews for the same items $j$ written by other users $k \ne i$. For a given user $i \in U$, we define the profile $P_i$ as the union of:
\begin{itemize}[topsep=0pt,itemsep=0pt,parsep=0pt,partopsep=0pt]
    \item the set of interactions authored by user $i$: $\{(i,j) \in E\}$,
    \item the set of interactions for the same items $j$ written by other users $k \ne i$: $\{(k,j) \in E \mid (i,j) \in E\}$.
\end{itemize}

\begin{align}
P_i &= \{(i,j) \in E\} \cup \{(k,j) \in E \;|\; (i,j) \in E\} \\
    &\forall j \in V,\; k \in U,\; k \ne i \nonumber
\end{align}

Due to context window limitations and efficiency considerations, we apply retrieval augmentation to select only the most relevant entries from $P_i$ for conditioning the model. Given an input sample $(x, y)$ for user $i$, the PGraphRAG workflow proceeds in three steps: a query function, a graph-based retrieval module, and a prompt construction function, as illustrated in Figure~\ref{fig:overview-personalized-graph-based-LLM}:

\begin{enumerate}[topsep=0pt,itemsep=2pt,parsep=0pt,partopsep=0pt]
    \item \textbf{Query Function (\( \phi_q \))}: The query function transforms the input $x$ into a query $q$ for retrieval.
    \item \textbf{Graph-Based Retrieval (\( \mathcal{R} \))}: The retrieval function \( \mathcal{R}(q, G, k) \) takes as input the query $q$, the bipartite graph $G$, and a threshold $k$. It first constructs the user profile $P_i$ from $G$ as defined above, and then retrieves the top-$k$ most relevant entries from the user profile $P_i$ with respect to $q$.
    \item \textbf{Prompt Construction (\( \phi_p \))}: The prompt construction assembles a personalized prompt for user $i$ by combining the input $x$ with the retrieved entries.
\end{enumerate}

The final input to the LLM is a personalized, context-augmented prompt $\tilde{x}$ defined as:
\begin{align}
\tilde{x} = \phi_p(x, \mathcal{R}(\phi_q(x), G, k))
\end{align}

The pair $(\tilde{x}, y)$ is then used for inference or fine-tuning. This modular pipeline enables efficient, graph-aware personalization across diverse tasks and user sparsity levels.

\noindent
\textbf{Modularity and Extensibility.}  
While we define $P_i$ as a hybrid of user-authored and neighbor-authored interactions, PGraphRAG is modular by design. The underlying graph can be leveraged in alternative ways depending on the application: for example, practitioners may define $P_i$ using only user-specific data, only neighbor interactions, or other graph-based traversal strategies (e.g., multi-hop reasoning or community-based filtering). Each component of the framework—query formulation, retrieval logic, and prompt construction—can be adapted independently, making PGraphRAG extensible to a wide range of personalized retrieval scenarios. In addition, the retrieval module supports plug-and-play compatibility with a variety of retrievers, such as BM25, or Contriever, allowing flexibility in balancing speed, semantic relevance, and computational cost.

%% file: text/exp-updated.tex
\section{Experiments}\label{sec:exp}

\paragraph{Setup.}
We evaluate our methods using two LLM backbones. The first is the LLaMA 3.1 8B Instruct model~\cite{touvron2023llamaopenefficientfoundation}, implemented with the Huggingface \texttt{transformers} library and configured to generate up to 512 tokens. The second is the GPT-4o-mini model~\cite{openai2024gpt4ocard}, accessed via the Azure OpenAI Service~\cite{azure_openai_gpt4o_mini} using the \texttt{AzureOpenAI} interface, with a decoding temperature of 0.4. All experiments are conducted on an NVIDIA A100 GPU with 80GB of memory.

\vspace{1mm}
\paragraph{Dataset Splits and Graph Construction}
We construct bipartite user-entity graphs and split users into training, development, and test sets while preserving connectivity. Full details on data construction, neighbor filtering, and stratification are provided in Appendix~\ref{sec:app_up}.

\vspace{1mm}
\noindent\textbf{Graph Construction. }
We construct a bipartite user-entity graph from the selected user profiles in the validation and test splits. Each user node is connected to entity nodes (e.g., products, hotels, feedback targets) based on authored content, with edges representing user interactions such as reviews, summaries, or ratings. This graph supports two retrieval configurations: (1) \textit{user-only}, which retrieves content authored solely by the target user (i.e., from their personal profile), and (2) \textit{user+neighbor}, which additionally includes content from neighboring users who have interacted with the shared target entity. In both modes, the retrieved content defines the personalized context passed to the language model.

\vspace{1mm}
\noindent\textbf{Ranking and Retrieval. }
The query used for retrieval varies by task type: for \emph{Long Text Generation}, we use the review title; for \emph{Short Text Generation}, the review text; and for \emph{Ordinal Classification}, a combination of title and text. We apply two retrieval models—BM25~\cite{bm25} and Contriever~\cite{lei-etal-2023-unsupervised} to select the top-$k$ ($k=5$) most relevant entries from either the user-only or user+neighbor profiles. To enforce consistency between users with high activity and cold-start users, we cap retrieval at $k$, even if more candidate entries are available (see Table~\ref{tab:review_distribution} and Figure~\ref{fig:profile_distribution}). All textual inputs are tokenized using NLTK’s \texttt{word\_tokenize}. We use the default settings for both retrieval models; for Contriever, mean pooling is applied over token embeddings.

\vspace{1mm}
\noindent\textbf{LLM Prompt Generation. }
Once the top-$k$ entries are retrieved, we construct a \emph{template-based prompt} that includes both the user’s query (e.g., a request for a full review, a title, or a rating) and the contextual information from the graph. This prompt is passed to the LLM for generation. An illustration of task-specific prompt formatting is shown in Figure~\ref{fig:task_prompts}.

\vspace{1mm}
\paragraph{Baseline Methods. }
We compare PGraphRAG against both non-personalized and personalized baselines.  
(1) \emph{No-Retrieval} constructs the prompt without any retrieval augmentation; the LLM generates the output solely from the query.  
(2) \emph{Random-Retrieval} augments the prompt with content randomly sampled from all user profiles, introducing unrelated context.  
(3) \emph{LaMP}~\cite{salemi2024lamp} is a personalized baseline that augments the prompt using content from the target user’s own history (e.g., previously written reviews).

\vspace{1mm}
\paragraph{Evaluation. }
We evaluate each method by providing task-specific inputs and comparing generated outputs against reference labels. For generation tasks (long and short text), we report ROUGE-1, ROUGE-L~\cite{lin-2004-rouge}, and METEOR~\cite{banerjee-lavie-2005-meteor} scores. For rating prediction tasks, we measure mean absolute error (MAE) and root mean squared error (RMSE).

\subsection{Baseline Comparison}

We compare PGraphRAG against baselines on the three task types in our benchmark --- long-text generation, short-text generation, and rating prediction. 
\input{tables/model-longtext-test}

\paragraph{Long Text Generation.}
Tables~\ref{tab:model-longtext-test} and~\ref{tab:model-longtext-dev} show that PGraphRAG consistently outperforms all baseline methods—including No-Retrieval, Random-Retrieval, and LaMP—across ROUGE-1, ROUGE-L, and METEOR metrics. The largest performance gains are observed in Task~\ttwo, where PGraphRAG achieves +32.1\% in ROUGE-1, +21.7\% in ROUGE-L, and +25.7\% in METEOR over the LaMP baseline using the \lla\ model. These improvements highlight the benefits of incorporating structured, graph-based context beyond user history. 

\paragraph{Short Text Generation.}
Tables~\ref{tab:model-shorttext-test} and~\ref{tab:model-shorttext-dev} show that PGraphRAG outperforms the baselines in most cases. In Task~\tfiv, PGraphRAG achieves consistent gains over LaMP in the \lla\ model: ROUGE-1 (+5.6\%), ROUGE-L (+5.9\%), and METEOR (+6.8\%). These improvements, while smaller than those in long-form tasks, reflect the limited headroom for personalization in very short text generation tasks such as review title. Because the target texts are extremely brief, minor lexical differences can significantly affect overlap-based metrics, and there are fewer opportunities for retrieved context to meaningfully influence generation.

\input{tables/model-shorttext-test}

\paragraph{Ordinal Classification.}
Tables~\ref{tab:model-rating-test} and~\ref{tab:model-rating-dev} show that PGraphRAG yields modest improvements over LaMP in rating prediction tasks. It outperforms LaMP in 1 out of 4 tasks with \lla\ and in 2 out of 4 tasks with GPT. The largest gains are observed on the Multilingual Product Ratings task, with improvements in MAE (+1.75\%) and RMSE (+1.12\%) for \lla, and MAE (+2.16\%) and RMSE (+3.17\%) for GPT. These gains, while small, suggest that user profiles can aid numerical prediction when meaningful variability exists across user preferences. In domains like hotel experiences or digital products, where user expectations tend to be homogeneous, graph-based personalization may offer limited additional signal.

\subsection{Ablation Studies}
We conduct ablation experiments to assess the impact of different retrieval configurations on PGraphRAG’s performance. Specifically, we vary the retrieval depth (i.e., top-$k$), the retrieval scope (user-only vs. user+neighbors), and the retriever model (BM25 vs. Contriever). Full results and analysis are provided in Appendix~\ref{sec:app_up}.

%% file: tables/model-longtext-test.tex
\begin{table*}[h!]
\centering
\renewcommand{\arraystretch}{0.85}
\setlength{\abovecaptionskip}{2pt} 
\setlength{\belowcaptionskip}{2pt} 
\setlength{\textfloatsep}{5pt}     
\begin{adjustbox}{width=\textwidth}
\begin{tabular}{llcccc}
\toprule
\textbf{Long Text Generation} & \textbf{Metric} & \makecell{\textbf{PGraphRAG}} & \textbf{\user} & \textbf{\none} & \textbf{\rand} \\
\midrule
\multicolumn{2}{l}{\textbf{\emph{\lla}}} \\
\midrule
\multirow{3}{*}{Task $1$: User-Product Review Generation}
& ROUGE-1 & \textbf{0.178} & 0.173 & 0.172 & 0.124 \\
& ROUGE-L & \textbf{0.129} & 0.129 & 0.123 & 0.094 \\
& METEOR  & 0.151 & 0.138 & \textbf{0.154} & 0.099 \\ 
\midrule
\multirow{3}{*}{Task $2$: Hotel Experiences Generation}
& ROUGE-1 & \textbf{0.263} & 0.199 & 0.231 & 0.216 \\
& ROUGE-L & \textbf{0.157} & 0.129 & 0.145 & 0.132 \\
& METEOR  & \textbf{0.191} & 0.152 & 0.153 & 0.152 \\ 
\midrule
\multirow{3}{*}{Task $3$: Stylized Feedback Generation}
& ROUGE-1 & \textbf{0.217} & 0.186 & 0.190 & 0.184 \\
& ROUGE-L & \textbf{0.158} & 0.134 & 0.131 & 0.108 \\
& METEOR  & \textbf{0.178} & 0.177 & 0.167 & 0.122 \\ 
\midrule
\multirow{3}{*}{Task $4$: Multilingual Product Review Generation}
& ROUGE-1 & \textbf{0.188} & 0.176 & 0.174 & 0.146 \\
& ROUGE-L & \textbf{0.147} & 0.141 & 0.136 & 0.116 \\
& METEOR  & \textbf{0.145} & 0.125 & 0.131 & 0.109 \\ 
\midrule
\multicolumn{2}{l}{\textbf{\emph{GPT-4o-mini}}} \\
\midrule
\multirow{3}{*}{Task $1$: User-Product Review Generation}
& ROUGE-1 & \textbf{0.189} & 0.171 & 0.169 & 0.159 \\
& ROUGE-L & \textbf{0.130} & 0.117 & 0.116 & 0.114 \\
& METEOR  & \textbf{0.196} & 0.176 & 0.177 & 0.153 \\ 
\midrule
\multirow{3}{*}{Task $2$: Hotel Experiences Generation}
& ROUGE-1 & \textbf{0.263} & 0.221 & 0.223 & 0.234 \\
& ROUGE-L & \textbf{0.152} & 0.135 & 0.135 & 0.139 \\
& METEOR  & \textbf{0.206} & 0.164 & 0.166 & 0.181 \\ 
\midrule
\multirow{3}{*}{Task $3$: Stylized Feedback Generation}
& ROUGE-1 & \textbf{0.211} & 0.185 & 0.187 & 0.177 \\
& ROUGE-L & \textbf{0.140} & 0.123 & 0.123 & 0.121 \\
& METEOR  & \textbf{0.202} & 0.183 & 0.189 & 0.165 \\ 
\midrule
\multirow{3}{*}{Task $4$: Multilingual Product Review Generation}
& ROUGE-1 & \textbf{0.194} & 0.168 & 0.170 & 0.175 \\
& ROUGE-L & \textbf{0.144} & 0.125 & 0.128 & 0.133 \\
& METEOR  & \textbf{0.171} & 0.154 & 0.152 & 0.149 \\ 
\bottomrule
\end{tabular}
\end{adjustbox}
\caption{
Zero-shot performance on the test set for the Long Text Generation tasks using \emph{LLaMA-3.1-8B-Instruct} and \emph{GPT-4o-mini}. For each model, the best retriever configuration was selected based on validation performance.
}
\label{tab:model-longtext-test}
\end{table*}

%% file: tables/model-shorttext-test.tex
\begin{table*}[h!]
\centering
\renewcommand{\arraystretch}{0.85}
\setlength{\abovecaptionskip}{2pt} 
\setlength{\belowcaptionskip}{2pt} 
\setlength{\textfloatsep}{5pt}     
\begin{adjustbox}{width=\textwidth}
\begin{tabular}{llcccc}
\toprule
\textbf{Short Text Generation} & \textbf{Metric} & \makecell{\textbf{PGraphRAG}} & \textbf{\user} & \textbf{\none} & \textbf{\rand} \\
\midrule
\multicolumn{2}{l}{\textbf{\emph{\lla}}} \\
\midrule
\multirow{3}{*}{Task $5$: \tfiv}
& ROUGE-1 & \textbf{0.131} & 0.124 & 0.121 & 0.103 \\
& ROUGE-L & \textbf{0.125} & 0.118 & 0.115 & 0.098 \\
& METEOR  & \textbf{0.125} & 0.117 & 0.112 & 0.096 \\
\midrule
\multirow{3}{*}{Task $6$: \tsix}
& ROUGE-1 & \textbf{0.127} & 0.126 & 0.122 & 0.118 \\
& ROUGE-L & \textbf{0.118} & 0.117 & 0.114 & 0.110 \\
& METEOR  & 0.102 & \textbf{0.106} & 0.101 & 0.093 \\
\midrule
 \multirow{3}{*}{Task $7$: \tsev}
& ROUGE-1 & \textbf{0.149} & 0.140 & 0.136 & 0.133 \\
& ROUGE-L & \textbf{0.142} & 0.134 & 0.131 & 0.123 \\
& METEOR  & \textbf{0.142} & 0.136 & 0.129 & 0.121 \\
\midrule
\multirow{3}{*}{Task $8$: \teig}
& ROUGE-1 & 0.124 & 0.121 & \textbf{0.125} & 0.120 \\
& ROUGE-L & 0.116 & \textbf{0.122} & 0.117 & 0.110 \\
& METEOR  & \textbf{0.108} & 0.094 & 0.092 & 0.103 \\
\midrule
\multicolumn{2}{l}{\textbf{\emph{\gpt}}} \\
\midrule
\multirow{3}{*}{Task $5$: \tfiv}
& ROUGE-1 & \textbf{0.115} & 0.108 & 0.113 & 0.102 \\
& ROUGE-L & \textbf{0.112} & 0.105 & 0.110 & 0.099 \\
& METEOR  & \textbf{0.099} & 0.091 & 0.093 & 0.085 \\
\midrule
\multirow{3}{*}{Task $6$: \tsix}
& ROUGE-1 & \textbf{0.116} & 0.108 & 0.114 & 0.112 \\
& ROUGE-L & \textbf{0.111} & 0.104 & 0.109 & 0.107 \\
& METEOR  & \textbf{0.081} & 0.075 & 0.079 & 0.076 \\
\midrule
\multirow{3}{*}{Task $7$: \tsev}
& ROUGE-1 & \textbf{0.122} & 0.113 & 0.114 & 0.115 \\
& ROUGE-L & \textbf{0.118} & 0.109 & 0.110 & 0.111 \\
& METEOR  & \textbf{0.104} & 0.096 & 0.097 & 0.093 \\
\midrule
\multirow{3}{*}{Task $8$: \teig}
& ROUGE-1 & 0.111 & 0.115 & \textbf{0.118} & 0.108 \\
& ROUGE-L & 0.105 & 0.107 & \textbf{0.110} & 0.102 \\
& METEOR  & 0.083 & 0.088 & \textbf{0.089} & 0.078 \\
\bottomrule
\end{tabular}
\end{adjustbox}
\caption{
Zero-shot performance on the test set for the Short Text Generation tasks using \emph{LLaMA-3.1-8B-Instruct} and \emph{GPT-4o-mini}. For each model, the best retriever configuration was selected based on validation performance.
}
\label{tab:model-shorttext-test}
\end{table*}

%% file: text/appendix-up.tex
\section{Appendix} \label{sec:app_up}

\subsection{Data Construction and Splitting}

To construct the user–item interaction graph, we represent users and domain-specific entities (e.g., products, hotels, feedback targets) as nodes, with edges corresponding to user-generated content (e.g., reviews, summaries, ratings). To support graph-based personalization, we require that each selected user has at least one interaction with an entity that is also associated with another user --- i.e., a shared neighbor in the bipartite graph. If a randomly selected user interaction does not meet this criterion, we instead sample a different interaction from the same profile. Users without any neighbor-compatible interactions remain in the dataset but are excluded from gold-label selection, since sampling is performed at the edge level rather than over full profiles. This filtering ensures that the graph remains connected and supports comparative evaluation and cold-start scenarios, where even users with minimal history share contextually linked entities with others.

After identifying each user's valid neighbor-linked interaction(s), we divide users into training, development, and test sets while preserving graph connectivity across splits. To ensure that personalization signals remain intact, we apply two levels of neighbor preservation:

\begin{enumerate}
    \item \textbf{Global Neighbor Preservation:} Entities with multiple associated users are grouped so that at least one other user in the same split has interacted with the same entity.
    \item \textbf{Local Neighbor Preservation:} Once a user is assigned to a split, any other users who interacted with the same entity are also placed in that split to maintain graph connectivity.
\end{enumerate}

We further stratify each split based on user profile size to match the original distribution of user activity while preserving both global and local connectivity. This joint control over profile stratification and neighbor assignment ensures that the resulting graphs in each split maintain realistic interaction patterns and structural properties. Graph statistics are shown in Table~\ref{tab:graph-stats}, task-level data statistics in Table~\ref{tab:task-stats}, and dataset splits in Table~\ref{tab:split-stats}.

\begin{table}[h!]
\centering
\small
\resizebox{\columnwidth}{!}{
\begin{tabular}{lccc}
\toprule
\textbf{Dataset} & \textbf{Train Size} & \textbf{Validation Size} & \textbf{Test Size} \\
\midrule
User-Product Review & 20,000 & 2,500 & 2,500 \\
Multilingual Product Review & 20,000 & 2,500 & 2,500 \\
Stylized Feedback & 20,000 & 2,500 & 2,500 \\
Hotel Experiences & 9,000 & 2,500 & 2,500 \\
\bottomrule
\end{tabular}}
\caption{Dataset split sizes across training, validation, and test sets for the four domains.}
\label{tab:split-stats}
\end{table}

\input{tables/gains_table}

\subsection{Performance Gains}

Table~\ref{tab:percent-gain} shows the relative percent gains of PGraphRAG compared to LaMP across Tasks 1--7. Notably, Task 8 (\teig) shows reduced gains, which we attribute to cultural differences in review conventions---for example, the frequent use of the generic phrase Muito bom'' (Very good'') in Brazilian Portuguese titles. In long-text generation with GPT-4o-mini, PGraphRAG achieves improvements of approximately 15\% in ROUGE-1, 13\% in ROUGE-L, and 15\% in METEOR. Similar trends are seen with LLaMA-3.1-8B, with improvements of 15\%, 11\%, and 13\% respectively. In short-text generation, GPT shows improvements of ~5\% across all metrics, while LLaMA gains range from 2--6\%.

In addition, Table \ref{tab:review_distribution} shows the review density per product, where sparsity is balanced from the original graph for both product and user nodes.

\begin{table}[h!]
\centering
\resizebox{\columnwidth}{!}{
\begin{tabular}{c c c}
\toprule
\textbf{Reviews ($n$)} & \textbf{Exact Count (Pct.)} & \textbf{Cumulative Count (Pct.)} \\
\midrule
1  & 25,530 (49.69\%)  & 25,530 (49.69\%) \\
2  & 9,488  (18.47\%)  & 35,018 (68.16\%) \\
3  & 4,784  ( 9.31\%)  & 39,802 (77.47\%) \\
4  & 2,639  ( 5.14\%)  & 42,441 (82.61\%) \\
5  & 1,836  ( 3.57\%)  & 44,277 (86.18\%) \\
\bottomrule
\end{tabular}}
\caption{Distribution of review counts per product in the Amazon Review dataset (Tasks 1, 5, and 9). Most products have five or fewer reviews.}
\label{tab:review_distribution}
\end{table}

\input{tables/model-rating-test}

\subsection{Prompt and Output Examples}

\begin{figure*}[htbp]
\centering
\includegraphics[width=\textwidth]{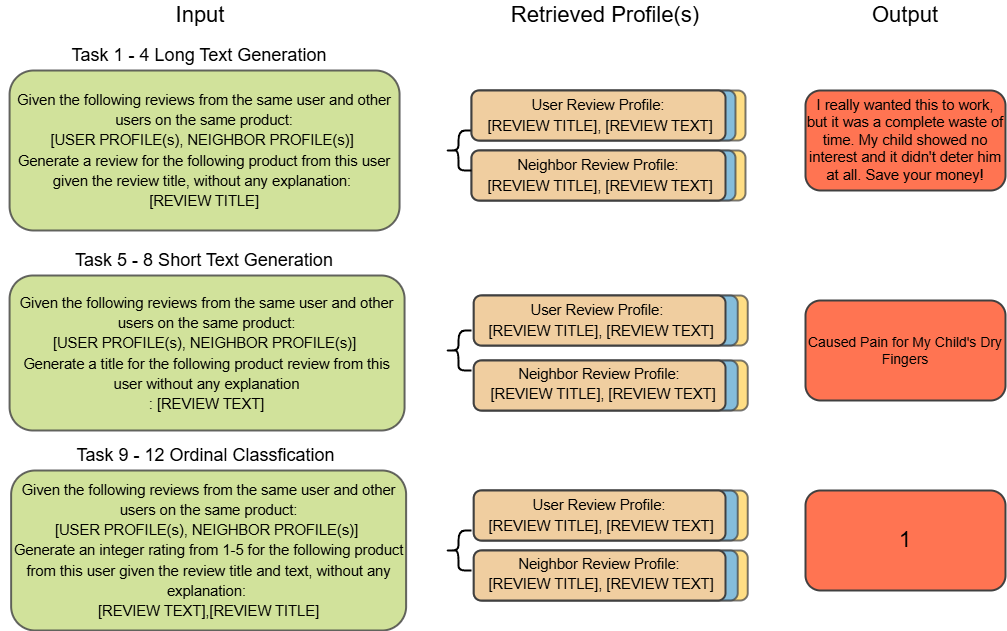}
\caption{Prompt configurations used for each task type. Teletype placeholders (e.g., \texttt{\{\{title\}\}}) are replaced with task-specific input and retrieved context at inference time.}
\label{fig:task_prompts}
\end{figure*}

Figure~\ref{fig:task_prompts} shows the prompt template across task types. Below, we show the output for Task~2 comparing PGraphRAG and LaMP against the gold label. PGraphRAG captures specific contextual cues from the graph (e.g., correct location and hotel amenities), while LaMP's output often relies on generic context from the target user's own reviews, which leads to generating irrelevant content (e.g., wrong hotel location). 

\begin{tcolorbox}[colback=blue!5, colframe=blue!40, sharp corners, boxrule=0.5pt, width=\linewidth,
title=\textbf{Gold Review}, fonttitle=\bfseries, fontupper=\small]
\textbf{Title:} Amazing stay! \
\textbf{Review:} Had a great stay, our room was very clean and very peaceful. It is in the heart of DT Seattle. We took the train to the hotel and it was pretty easy. We enjoyed walking all over downtown Seattle\textemdash no need to rent a car. Major shopping and amazing restaurants right around the corner! We will stay here again!
\end{tcolorbox}

\begin{tcolorbox}[colback=orange!5, colframe=orange!40, sharp corners, boxrule=0.5pt, width=\linewidth,
title=\textbf{LaMP}, fonttitle=\bfseries, fontupper=\small]
\textbf{Title:} Amazing stay! \
\textbf{Review:} I recently had an amazing stay at the Newark Garden Hotel. Despite the circumstances that brought my family together, the hotel provided a comforting and welcoming environment. The rooms were immaculate, and the staff went above and beyond to ensure we felt at home. The hot breakfast each morning was a delightful touch that made our mornings brighter. I highly recommend this hotel for anyone looking for a pleasant and accommodating stay!
\end{tcolorbox}

\begin{tcolorbox}[colback=green!5, colframe=green!40, sharp corners, boxrule=0.5pt, width=\linewidth,
title=\textbf{PGraphRAG}, fonttitle=\bfseries, fontupper=\small]
\textbf{Title:} Amazing stay! \
\textbf{Review:} I had an amazing stay at the Grand Hyatt Seattle! From the moment I checked in, the staff was incredibly welcoming and attentive. The room was spacious, clean, and beautifully appointed, with a stunning view of the city. The bathroom was a highlight, featuring a luxurious soaking tub and a large walk-in shower. The location was perfect, just a short walk to Pike Place Market and other attractions. I also enjoyed the fantastic amenities, including the well-equipped gym and the delicious food options available at the hotel. Overall, it was a wonderful experience, and I can't wait to return!
\end{tcolorbox}

\subsection{PGraphRAG Ablation Details}
\label{sec:appendix_ablation_pgrag}

\input{tables/merged_pgrag_ablation_long}
\input{tables/merged_pgrag_ablation_short}

To assess the contributions of user-specific and neighbor-derived context in our retrieval framework, we conduct an ablation study comparing three variants of PGraphRAG:

\begin{itemize}
\item \textbf{PGraphRAG}: The full method, which retrieves context from both the target user’s profile and neighboring users who share entities (e.g., items or experiences).
\item \textbf{PGraphRAG-N}: A neighbor-only variant that excludes the target user’s own interactions and relies solely on neighboring users for context.
\item \textbf{PGraphRAG-U}: A user-only variant that restricts retrieval to the target user’s own history, ignoring all neighbor signals.
\end{itemize}

Table~\ref{tab:merged_pgrag_ablation_long} shows the results for long-text generation (Tasks 1–4) using GPT-4o-mini and LLaMA-3.1-8B. Both PGraphRAG and PGraphRAG-N consistently outperform PGraphRAG-U across datasets, highlighting the value of graph-based retrieval. Notably, PGraphRAG-N performs on par with or slightly below the full PGraphRAG method, suggesting that neighboring-user context alone is often sufficient for high-quality personalization --- especially in low-profile or cold-start scenarios where the target user's history is sparse.

Results for short-text generation tasks (Tasks 5–8) are shown in Table~\ref{tab:merged_pgrag_ablation_short}. Similar patterns hold, with PGraphRAG and PGraphRAG-N outperforming PGraphRAG-U across most tasks. One exception is Task \tsix, where PGraphRAG-U slightly outperforms all graph-based variants, possibly due to limited variation in the data or a mismatch between neighbor context and task-specific semantics.

\subsection{Impact of the Retrieved Items $k$}
\label{sec:appendix_ablation_k}

To understand how the size of the retrieved context affects performance, we conduct an ablation study varying the number of retrieved entries $k \in {1, 2, 4}$. Table~\ref{tab:merged_k_long} reports results for long-text generation (Tasks 1–4), using \textit{GPT-4o-mini} and \textit{LLaMA-3.1-8B-Instruct}. Corresponding results for short-text generation (Tasks 5–8) appear in Table~\ref{tab:merged_k_short}.

Overall, increasing $k$ generally leads to improved generation performance across tasks and models. This trend highlights the value of larger retrieved contexts, which provide richer signals about user preferences and item semantics. The gains are especially evident when moving from $k=1$ to $k=2$, though marginal returns diminish between $k=2$ and $k=4$ in some cases.

That said, the benefit of higher $k$ values is constrained by data sparsity. Many user profiles contain fewer than four qualifying interactions—especially in cold-start settings. In such cases, the retriever returns all available entries, even if they are fewer than the specified $k$. As a result, the effective retrieved context size varies across users, especially in the low-profile regime. This behavior reflects the practical limitations of personalization at scale and underscores the importance of designing retrieval-aware systems that can operate under sparse supervision.

\input{tables/merged_k_long}
\input{tables/merged_k_short}

\subsection{Impact of Retriever Method $\mathcal{R}$}
\label{sec:appendix_ablation_retriever}

We evaluate how the choice of retriever affects the performance of PGraphRAG by comparing two retrieval backends: BM25, a sparse keyword-based retriever, and Contriever, a dense unsupervised retriever based on sentence embeddings.

Table~\ref{tab:merged_ret_long} reports results for long-text generation (Tasks 1–4), and Table~\ref{tab:merged_ret_short} provides results for short-text generation (Tasks 5–8). Across both GPT-4o-mini and LLaMA-3.1-8B-Instruct models, we observe that PGraphRAG performs consistently well regardless of the retrieval method. The differences between BM25 and Contriever are minor, and no retriever dominates across all datasets or metrics.

These findings indicate that PGraphRAG is robust to the choice of retriever and does not rely on fine-tuned or heavily engineered retrieval strategies. While BM25 sometimes yields slightly higher scores, the overall parity suggests that our graph-based retrieval and prompting framework can effectively integrate contextual signals from either sparse or dense retrieval methods.

\FloatBarrier
\input{tables/merged_ret_long}
\input{tables/merged_ret_short}

\subsection{Impact of Ranked Retrieval}
\label{sec:appendix_ranked_retrieval}

\input{tables/gpt-norank-shortext}

Table~\ref{tab:gpt-norank-text} evaluates the role of ranking in PGraphRAG by comparing the following retrieval variants:

\begin{enumerate}
    \item PGraphRAG*: retrieves $k=4$ randomly sampled entries from the profile without ranking.
    \item PGraphRAG**: retrieves and includes all available context within the model’s input limit (i.e., $k \rightarrow \infty$).
\end{enumerate}

As expected, PGraphRAG** performs best due to its access to a larger and more diverse context. However, our focus is on the impact of removing ranking while keeping $k$ fixed.

Removing ranking (PGraphRAG $\rightarrow$ PGraphRAG*) leads to a drop in ROUGE-1 of 2.29\% for long-text generation and 3.18\% for short-text tasks. The effect is also visible in user-only retrieval (PGraphRAG-U $\rightarrow$ PGraphRAG-U*), with decreases of 0.92\% and 1.98\% for long- and short-text tasks, respectively. These consistent declines underscore the importance of ranking in identifying relevant context.

While PGraphRAG** demonstrates the upper bound of performance, its scalability is limited due to cost and context length constraints. In contrast, ranked retrieval with a fixed $k$ (as in PGraphRAG) offers a strong balance between performance and efficiency, making it more suitable for real-world deployment.

\subsection{Evaluating Different GPT Variants}
To compare the performance of different GPT variants, we evaluate PGraphRAG using a fixed retrieval configuration (BM25, $k = 4$) across two OpenAI models: GPT-4o-mini and GPT-o1. Among these, \gpt{} demonstrated the best trade-off between accuracy, cost, and consistency on long-text generation tasks. 

\begin{figure}[htbp]
    \centering
    \includegraphics[width=\columnwidth,keepaspectratio]{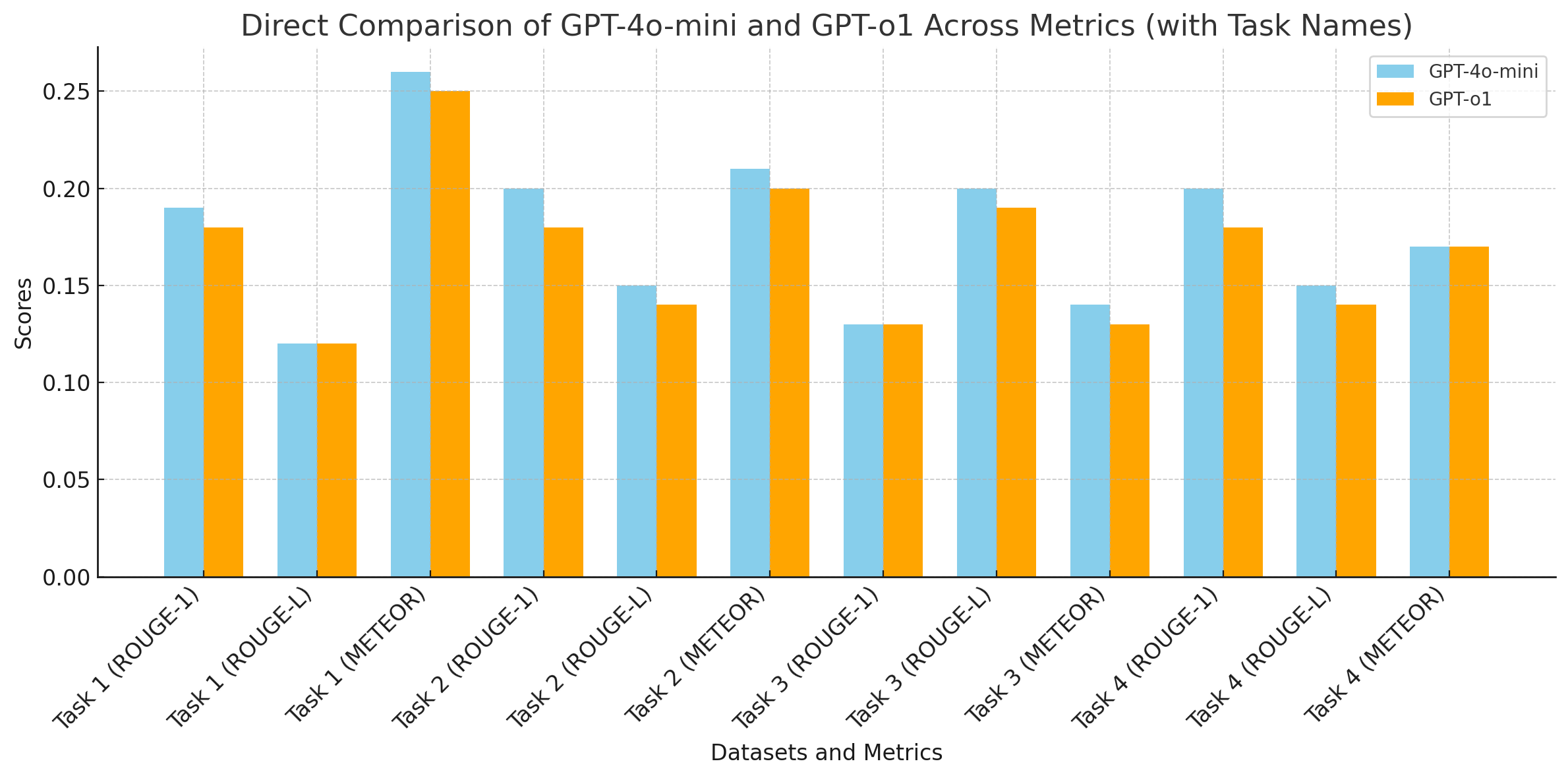}
    \caption{Comparison of \textit{GPT-4o-mini} and \textit{GPT-o1-preview} on the test set across Tasks 1--4 using BM25 retriever with $k=4$.}
    \label{fig:model_comparison}
\end{figure}

\subsection{Impact of Length Constraints in GPT Model}
In short-text generation tasks, controlling output length is essential to balance informativeness and conciseness. We evaluate the effect of fixed output constraints of 3, 5, and 10 words. Empirically, a 5-word constraint offers the best trade-off across evaluation metrics, yielding higher-quality outputs with minimal verbosity. We therefore adopt 5-word outputs as the default setting for all short-text generation experiments.

\begin{figure}[htbp]
    \centering
    \includegraphics[width=\columnwidth]{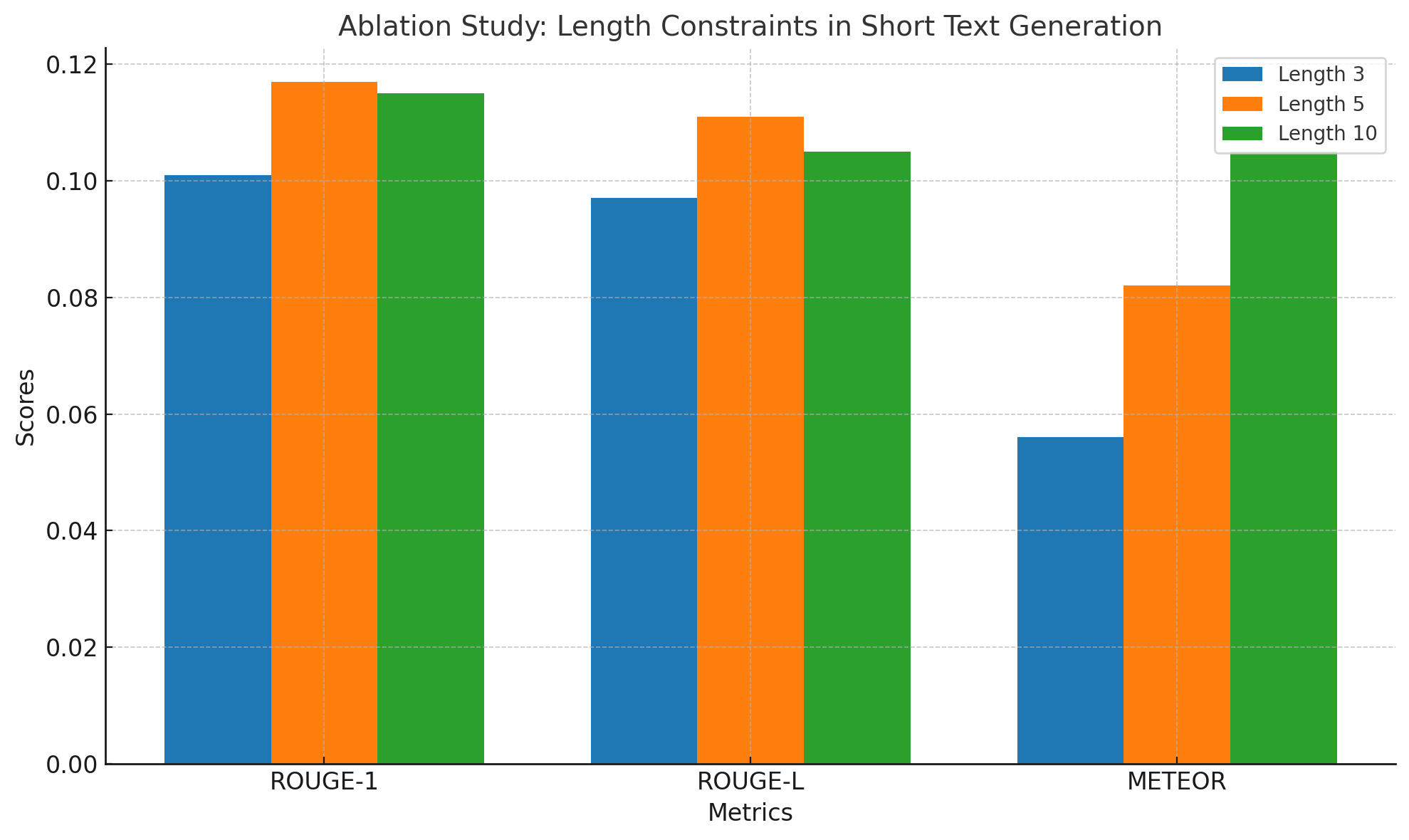}
    \caption{Effect of different output length constraints (3, 5, and 10 words) on short-text generation performance using PGraphRAG, measured on the validation set.}
    \label{fig:length_ablation}
\end{figure}

\subsection{Validation Results}

We conduct extensive validation experiments across all representative tasks, evaluating all combinations of language models, retrieval strategies, and top-$k$ settings. The goal is to identify the most effective configuration for each task prior to test-time evaluation.

Results are reported in Tables~\ref{tab:model-longtext-dev}, \ref{tab:model-shorttext-dev}, and \ref{tab:model-rating-dev}, corresponding to long-text generation, short-text generation, and ordinal classification tasks, respectively.

For each task, we select the best-performing configuration based on validation performance. These selected settings are then used in the test set evaluation. Notably, trends observed in the validation phase remain consistent in the test results, reinforcing the robustness of our setup.

\input{tables/model-longtext-dev}
\input{tables/model-shorttext-dev}
\input{tables/model-rating-dev}

\input{text/related-work-up}

%% file: tables/gains_table.tex
\begin{table*}[t]
    \centering
    \renewcommand{\arraystretch}{1.1}
    \resizebox{\textwidth}{!}{
    \begin{tabular}{llcccccccc}
        \toprule
        Model & Metric & Task 1 & Task 2 & Task 3 & Task 4 & Task 5 & Task 6 & Task 7 & Task 8 \\
        \midrule
        \multirow{3}{*}{\textbf{\textit{\gpt}}} 
        & ROUGE-1  & 10.53 & 18.96 & 14.05 & 15.48 & 6.48  & 7.41  & 7.96  & -3.48 \\
        & ROUGE-L  & 11.11 & 12.59 & 13.82 & 15.20 & 6.67  & 6.73  & 8.26  & -1.87 \\
        & METEOR   & 11.36 & 25.61 & 10.38 & 11.04 & 8.79  & 8.00  & 8.33  & -5.68 \\
        \midrule
        \multirow{3}{*}{\textbf{\textit{\lla}}} 
        & ROUGE-1  & 2.89  & 32.16 & 16.67 & 6.82  & 5.65  & 0.79  & 6.43  & 2.48  \\
        & ROUGE-L  & 0.00  & 21.71 & 17.91 & 4.26  & 5.93  & 0.85  & 5.97  & -4.92 \\
        & METEOR   & 9.42  & 25.66 & 0.56  & 16.00 & 6.84  & -3.77 & 4.41  & 14.89 \\
        \bottomrule
    \end{tabular}

    }
    \caption{Relative percentage gains of PGraphRAG over LaMP across Tasks 1--8 using \emph{GPT-4o-mini} and \emph{LLaMA-3.1-8B-Instruct}.}

    \label{tab:percent-gain}
\end{table*}

%% file: tables/model-rating-test.tex
\begin{table*}[h!]
\centering
\begin{adjustbox}{width=\textwidth}
\begin{tabular}{llcccc}
\toprule
\textbf{Ordinal Classfication} & \textbf{Metric} & \textbf{PGraphRAG} & \textbf{LaMP} & \textbf{No-retrieval} & \textbf{Random-retrieval} \\
\midrule
\multicolumn{6}{l}{\textbf{\textit{\lla}}} \\
\midrule
\multirow{2}{*}{Task 9: User Product Review Ratings}
    & MAE \textdownarrow  & 0.3400  & \textbf{0.3132}  & 0.3212  & 0.3272  \\
    & RMSE \textdownarrow & 0.7668  & \textbf{0.7230}  & 0.7313  & 0.7616  \\
\midrule
\multirow{2}{*}{Task 10: Hotel Experience Ratings}
    & MAE \textdownarrow  & 0.3688  & 0.3492  & \textbf{0.3340}  & 0.3804  \\
    & RMSE \textdownarrow & 0.6771  & 0.6527  & \textbf{0.6372}  & 0.6971  \\
\midrule
\multirow{2}{*}{Task 11: Stylized Feedback Ratings}
    & MAE \textdownarrow  & 0.3476  & \textbf{0.3268}  & 0.3256  & 0.3704  \\
    & RMSE \textdownarrow & 0.7247  & \textbf{0.6803}  & 0.6806  & 0.7849  \\
\midrule
\multirow{2}{*}{Task 12: Multi-lingual Product Ratings}
    & MAE \textdownarrow  & \textbf{0.4928}  & 0.5016  & 0.5084  & 0.5096  \\
    & RMSE \textdownarrow & \textbf{0.8367}  & 0.8462  & 0.8628  & 0.8542  \\
\midrule
\multicolumn{6}{l}{\textbf{\textit{\gpt}}} \\
\midrule
\multirow{2}{*}{Task 9: User Product Review Ratings}
    & MAE \textdownarrow  & 0.3832  & 0.3480  & \textbf{0.3448}  & 0.4188  \\
    & RMSE \textdownarrow & 0.7392  & \textbf{0.7065}  & \textbf{0.7065}  & 0.8082  \\
\midrule
\multirow{2}{*}{Task 10: Hotel Experience Ratings}
    & MAE \textdownarrow  & \textbf{0.3284}  & 0.3336  & 0.3336  & 0.3524  \\
    & RMSE \textdownarrow & \textbf{0.6083}  & 0.6197  & 0.6197  & 0.6384  \\
\midrule
\multirow{2}{*}{Task 11: Stylized Feedback Ratings}
    & MAE \textdownarrow  & 0.3476  & \textbf{0.3448}  & 0.3416  & 0.4080  \\
    & RMSE \textdownarrow & 0.6738  & \textbf{0.6669}  & 0.6711  & 0.7370  \\
\midrule
\multirow{2}{*}{Task 12: Multi-lingual Product Ratings}
    & MAE \textdownarrow  & \textbf{0.4348}  & 0.4444  & 0.4564  & 0.4700  \\
    & RMSE \textdownarrow & \textbf{0.7367}  & 0.7608  & 0.7718  & 0.8112  \\
\bottomrule
\end{tabular}
\end{adjustbox}
\caption{Performance comparison on rating prediction tasks (Tasks 9-12) using \emph{GPT-4o-mini} and \emph{LLaMA-3.1-8B}. }
\label{tab:model-rating-test}
\end{table*}

%% file: tables/merged_pgrag_ablation_long.tex
\begin{table*}[h!]
\centering
\renewcommand{\arraystretch}{0.85}
\captionsetup{skip=4pt}
\small
\begin{adjustbox}{width=\textwidth}
\begin{tabular}{llcccc}
\toprule
\textbf{Long Text Generation} & \textbf{Metric} & \textbf{PGraphRAG} & \textbf{PGraphRAG-N} & \textbf{PGraphRAG-U} \\
\midrule
\multicolumn{2}{l}{\textbf{\emph{\lla}}} \\
\midrule
\multirow{3}{*}{Task $1$: User-Product Review Generation}
& ROUGE-1 & 0.173 & \textbf{0.177} & 0.168 \\
& ROUGE-L & 0.124 & \textbf{0.127} & 0.125 \\
& METEOR  & 0.150 & \textbf{0.154} & 0.134 \\ 
\midrule
\multirow{3}{*}{Task $2$: Hotel Experiences Generation}
& ROUGE-1 & 0.263 & \textbf{0.272} & 0.197 \\
& ROUGE-L & 0.156 & \textbf{0.162} & 0.128 \\
& METEOR  & 0.191 & \textbf{0.195} & 0.121 \\ 
\midrule
\multirow{3}{*}{Task $3$: Stylized Feedback Generation}
& ROUGE-1 & \textbf{0.226} & 0.222 & 0.181 \\
& ROUGE-L & \textbf{0.171} & 0.165 & 0.134 \\
& METEOR  & \textbf{0.192} & 0.186 & 0.147 \\ 
\midrule
\multirow{3}{*}{Task $4$: Multilingual Product Review Generation}
& ROUGE-1 & \textbf{0.174} & 0.172 & 0.174 \\
& ROUGE-L & 0.139 & 0.137 & \textbf{0.141} \\
& METEOR  & \textbf{0.133} & 0.126 & 0.125 \\ 
\midrule
\multicolumn{2}{l}{\textbf{\emph{GPT-4o-mini}}} \\
\midrule
\multirow{3}{*}{Task $1$: User-Product Review Generation}
& ROUGE-1 & \textbf{0.186} & 0.185 & 0.169 \\
& ROUGE-L & \textbf{0.126} & 0.125 & 0.114 \\
& METEOR  & \textbf{0.187} & 0.185 & 0.170 \\ 
\midrule
\multirow{3}{*}{Task $2$: Hotel Experiences Generation}
& ROUGE-1 & 0.265 & \textbf{0.268} & 0.217 \\
& ROUGE-L & 0.152 & \textbf{0.153} & 0.132 \\
& METEOR  & 0.206 & \textbf{0.209} & 0.161 \\ 
\midrule
\multirow{3}{*}{Task $3$: Stylized Feedback Generation}
& ROUGE-1 & \textbf{0.205} & 0.204 & 0.178 \\
& ROUGE-L & \textbf{0.139} & 0.138 & 0.121 \\
& METEOR  & \textbf{0.203} & 0.198 & 0.178 \\ 
\midrule
\multirow{3}{*}{Task $4$: Multilingual Product Review Generation}
& ROUGE-1 & \textbf{0.191} & 0.190 & 0.164 \\
& ROUGE-L & \textbf{0.142} & 0.140 & 0.123 \\
& METEOR  & \textbf{0.173} & 0.169 & 0.155 \\ 
\bottomrule
\end{tabular}
\end{adjustbox}
\caption{Ablation study results for long text generation tasks using \emph{LLaMA-3.1-8B-Instruct} and \emph{GPT-4o-mini}. PGraphRAG-N represents Neighbors-only context retrieval and PGraphRAG-U represents User-only context retrieval.}
\label{tab:merged_pgrag_ablation_long}
\end{table*}

%% file: tables/merged_pgrag_ablation_short.tex
\begin{table*}[h!]
\centering
\renewcommand{\arraystretch}{0.85}
\captionsetup{skip=4pt}
\small
\begin{adjustbox}{width=\textwidth}
\begin{tabular}{llcccc}
\toprule
\textbf{Short Text Generation} & \textbf{Metric} & \textbf{PGraphRAG} & \textbf{PGraphRAG-N} & \textbf{PGraphRAG-U} \\
\midrule
\multicolumn{2}{l}{\textbf{\emph{\lla}}} \\
\midrule
\multirow{3}{*}{Task $5$: User Product Review Title Generation}
& ROUGE-1 & 0.125 & \textbf{0.129} & 0.115 \\
& ROUGE-L & 0.119 & \textbf{0.123} & 0.109 \\
& METEOR  & 0.117 & \textbf{0.120} & 0.111 \\ 
\midrule
\multirow{3}{*}{Task $6$: Hotel Experience Summary Generation}
& ROUGE-1 & 0.121 & \textbf{0.124} & 0.119 \\
& ROUGE-L & 0.113 & \textbf{0.115} & 0.111 \\
& METEOR  & 0.099 & 0.103 & \textbf{0.105} \\ 
\midrule
\multirow{3}{*}{Task $7$: Stylized Feedback Title Generation}
& ROUGE-1 & 0.132 & \textbf{0.135} & 0.128 \\
& ROUGE-L & 0.128 & \textbf{0.130} & 0.124 \\
& METEOR  & 0.129 & \textbf{0.132} & 0.124 \\ 
\midrule
\multirow{3}{*}{Task $8$: Multi-lingual Product Review Title Generation}
& ROUGE-1 & \textbf{0.131} & 0.131 & 0.124 \\
& ROUGE-L & \textbf{0.123} & 0.122 & 0.114 \\
& METEOR  & \textbf{0.118} & 0.110 & 0.098 \\ 
\midrule
\multicolumn{2}{l}{\textbf{\emph{GPT-4o-mini}}} \\
\midrule
\multirow{3}{*}{Task $5$: User Product Review Title Generation}
& ROUGE-1 & 0.111 & \textbf{0.116} & 0.112 \\
& ROUGE-L & 0.106 & \textbf{0.111} & 0.108 \\
& METEOR  & 0.097 & \textbf{0.099} & 0.095 \\ 
\midrule
\multirow{3}{*}{Task $6$: Hotel Experience Summary Generation}
& ROUGE-1 & 0.118 & \textbf{0.119} & 0.109 \\
& ROUGE-L & 0.112 & \textbf{0.113} & 0.104 \\
& METEOR  & \textbf{0.085} & \textbf{0.085} & 0.077 \\ 
\midrule
\multirow{3}{*}{Task $7$: Stylized Feedback Title Generation}
& ROUGE-1 & \textbf{0.109} & 0.107 & 0.108 \\
& ROUGE-L & \textbf{0.107} & 0.105 & 0.104 \\
& METEOR  & \textbf{0.096} & 0.094 & 0.091 \\ 
\midrule
\multirow{3}{*}{Task $8$: Multi-lingual Product Review Title Generation}
& ROUGE-1 & 0.108 & 0.109 & \textbf{0.116} \\
& ROUGE-L & 0.104 & 0.104 & \textbf{0.109} \\
& METEOR  & 0.082 & 0.089 & \textbf{0.091} \\ 
\bottomrule
\end{tabular}
\end{adjustbox}
\caption{Ablation study results for short text generation tasks using \emph{LLaMA-3.1-8B-Instruct} and \emph{GPT-4o-mini}. PGraphRAG-N represents Neighbors-only context retrieval and PGraphRAG-U represents User-only context retrieval.}
\label{tab:merged_pgrag_ablation_short}
\end{table*}

%% file: tables/merged_k_long.tex
\begin{table}[h]
\centering
\scriptsize 
\renewcommand{\arraystretch}{1.1} 
\resizebox{\columnwidth}{!}{ 
\begin{tabular}{llccc}
\toprule
\textbf{Long Text Generation} & \textbf{Metric} & $k=1$ & $k=2$ & $k=4$ \\
\midrule
\multicolumn{2}{l}{\textbf{\emph{\lla}}} \\
\midrule
\multirow{3}{*}{\makecell[l]{Task $1$: User-Product \\ Review Generation}} & ROUGE-1 & 0.160 & 0.169 & \textbf{0.173} \\
                           & ROUGE-L & 0.121 & \textbf{0.125} & 0.124 \\
                           & METEOR  & 0.125 & 0.138 & \textbf{0.150} \\ 
\midrule
\multirow{3}{*}{\makecell[l]{Task $2$: Hotel \\ Experiences Generation}} & ROUGE-1 & 0.230 & 0.251 & \textbf{0.263} \\
                           & ROUGE-L & 0.141 & 0.151 & \textbf{0.156} \\
                           & METEOR  & 0.152 & 0.174 & \textbf{0.191} \\ 
\midrule
\multirow{3}{*}{\makecell[l]{Task $3$: Stylized \\ Feedback Generation}} & ROUGE-1 & 0.200 & 0.214 & \textbf{0.226} \\
                           & ROUGE-L & 0.158 & 0.165 & \textbf{0.171} \\
                           & METEOR  & 0.154 & 0.171 & \textbf{0.192} \\ 
\midrule
\multirow{3}{*}{\makecell[l]{Task $4$: Multilingual \\ Product Review Generation}} & ROUGE-1 & 0.163 & 0.169 & \textbf{0.174} \\
                           & ROUGE-L & 0.134 & 0.137 & \textbf{0.139} \\
                           & METEOR  & 0.113 & 0.122 & \textbf{0.133} \\ 
\midrule
\multicolumn{2}{l}{\textbf{\emph{GPT-4o-mini}}} \\
\midrule
\multirow{3}{*}{\makecell[l]{Task $1$: User-Product \\ Review Generation}} & ROUGE-1 & 0.176 & 0.184 & \textbf{0.186} \\
                           & ROUGE-L & 0.121 & 0.125 & \textbf{0.126} \\
                           & METEOR  & 0.168 & 0.180 & \textbf{0.187} \\ 
\midrule
\multirow{3}{*}{\makecell[l]{Task $2$: Hotel \\ Experiences Generation}} & ROUGE-1 & 0.250 & 0.260 & \textbf{0.265} \\
                           & ROUGE-L & 0.146 & 0.150 & \textbf{0.152} \\
                           & METEOR  & 0.188 & 0.198 & \textbf{0.206} \\ 
\midrule
\multirow{3}{*}{\makecell[l]{Task $3$: Stylized \\ Feedback Generation}} & ROUGE-1 & 0.196 & 0.200 & \textbf{0.205} \\
                           & ROUGE-L & 0.136 & 0.136 & \textbf{0.139} \\
                           & METEOR  & 0.186 & 0.192 & \textbf{0.203} \\ 
\midrule
\multirow{3}{*}{\makecell[l]{Task $4$: Multilingual \\ Product Review Generation}} & ROUGE-1 & 0.163 & 0.169 & \textbf{0.174} \\
                           & ROUGE-L & 0.134 & 0.137 & \textbf{0.139} \\
                           & METEOR  & 0.113 & 0.122 & \textbf{0.133} \\ 
\bottomrule
\end{tabular}
}
\caption{Ablation study results showing the impact of varying $k$ (number of retrieved neighbors) on PGraphRAG's performance. Results are reported for \emph{LLaMA-3.1-8B-Instruct} and \emph{GPT-4o-mini} on long-text generation tasks (Tasks 1 - 4).}
\label{tab:merged_k_long}
\end{table}

%% file: tables/merged_k_short.tex
\begin{table}[h]
\centering
\scriptsize 
\renewcommand{\arraystretch}{1.1} 
\resizebox{\columnwidth}{!}{ 
\begin{tabular}{llccc}
\toprule
\textbf{Short Text Generation} & \textbf{Metric} & $k=1$ & $k=2$ & $k=4$ \\
\midrule
\multicolumn{2}{l}{\textbf{\emph{\lla}}} \\
\midrule
\multirow{3}{*}{\makecell[l]{Task $5$: User Product \\ Review Title Generation}} 
               & ROUGE-1   & \textbf{0.128} & 0.123 & 0.125 \\
               & ROUGE-L   & \textbf{0.121} & 0.118 & 0.119 \\
               & METEOR    & \textbf{0.123} & 0.118 & 0.117 \\
\midrule
\multirow{3}{*}{\makecell[l]{Task $6$: Hotel Experience \\ Summary Generation}} 
               & ROUGE-1   & \textbf{0.122} & 0.121 & 0.121 \\
               & ROUGE-L   & 0.112 & \textbf{0.114} & 0.113 \\
               & METEOR    & \textbf{0.104} & 0.102 & 0.099 \\
\midrule
\multirow{3}{*}{\makecell[l]{Task $7$: Stylized Feedback \\ Title Generation}} 
               & ROUGE-1   & 0.129 & \textbf{0.132} & \textbf{0.132} \\
               & ROUGE-L   & 0.124 & 0.126 & \textbf{0.128} \\
               & METEOR    & 0.129 & \textbf{0.130} & 0.129 \\
\midrule
\multirow{3}{*}{\makecell[l]{Task $8$: Multi-lingual Product \\ Review Title Generation}} 
               & ROUGE-1   & 0.129 & 0.126 & \textbf{0.131} \\
               & ROUGE-L   & 0.120 & 0.119 & \textbf{0.123} \\
               & METEOR    & 0.117 & 0.116 & \textbf{0.118} \\
\midrule
\multicolumn{2}{l}{\textbf{\emph{GPT-4o-mini}}} \\
\midrule
\multirow{3}{*}{\makecell[l]{Task $5$: User Product \\ Review Title Generation}} 
               & ROUGE-1   & \textbf{0.111} & 0.110 & \textbf{0.111} \\
               & ROUGE-L   & \textbf{0.106} & 0.105 & \textbf{0.106} \\
               & METEOR    & 0.093 & 0.094 & \textbf{0.097} \\
\midrule
\multirow{3}{*}{\makecell[l]{Task $6$: Hotel Experience \\ Summary Generation}} 
               & ROUGE-1   & 0.114 & 0.114 & \textbf{0.118} \\
               & ROUGE-L   & 0.109 & 0.109 & \textbf{0.112} \\
               & METEOR    & 0.082 & 0.082 & \textbf{0.085} \\
\midrule
\multirow{3}{*}{\makecell[l]{Task $7$: Stylized Feedback \\ Title Generation}} 
               & ROUGE-1   & 0.100 & 0.103 & \textbf{0.109} \\
               & ROUGE-L   & 0.098 & 0.101 & \textbf{0.107} \\
               & METEOR    & 0.087 & 0.090 & \textbf{0.096} \\
\midrule
\multirow{3}{*}{\makecell[l]{Task $8$: Multi-lingual Product \\ Review Title Generation}} 
               & ROUGE-1   & 0.104 & 0.104 & \textbf{0.108} \\
               & ROUGE-L   & 0.098 & 0.098 & \textbf{0.104} \\
               & METEOR    & 0.077 & 0.078 & \textbf{0.082} \\
\bottomrule
\end{tabular}
}
\caption{Ablation study results showing the impact of varying $k$ (number of retrieved neighbors) on PGraphRAG's performance. Results are reported for \emph{LLaMA-3.1-8B-Instruct} and \emph{GPT-4o-mini} on short-text generation tasks (Tasks 5-8).}
\label{tab:merged_k_short}
\end{table}

%% file: tables/merged_ret_long.tex
\begin{table}[ht]
\centering
\scriptsize 
\renewcommand{\arraystretch}{1.1} 
\resizebox{\columnwidth}{!}{ 
\begin{tabular}{llcc}
\toprule
\textbf{Long Text Generation} & \textbf{Metric} & \textbf{Contriever} & \textbf{BM25} \\
\midrule
\multicolumn{2}{l}{\textbf{\emph{\lla}}} \\
\midrule
\multirow{3}{*}{\makecell[l]{Task $1$: User-Product \\ Review Generation}}
               & ROUGE-1   & 0.172 & \textbf{0.173} \\
               & ROUGE-L   & 0.122 & \textbf{0.124} \\
               & METEOR    & \textbf{0.153} & 0.150 \\
\midrule
\multirow{3}{*}{\makecell[l]{Task $2$: Hotel \\ Experiences Generation}}
               & ROUGE-1   & 0.262 & \textbf{0.263} \\
               & ROUGE-L   & 0.155 & \textbf{0.156} \\
               & METEOR    & 0.190 & \textbf{0.191} \\
\midrule
\multirow{3}{*}{\makecell[l]{Task $3$: Stylized \\ Feedback Generation}}
               & ROUGE-1   & 0.195 & \textbf{0.226} \\
               & ROUGE-L   & 0.138 & \textbf{0.171} \\
               & METEOR    & 0.180 & \textbf{0.192} \\
\midrule
\multirow{3}{*}{\makecell[l]{Task $4$: Multilingual \\ Product Review Generation}}
               & ROUGE-1   & 0.172 & \textbf{0.174} \\
               & ROUGE-L   & 0.134 & \textbf{0.139} \\
               & METEOR    & \textbf{0.135} & 0.133 \\
\midrule
\multicolumn{2}{l}{\textbf{\emph{GPT-4o-mini}}} \\
\midrule
\multirow{3}{*}{\makecell[l]{Task $1$: User-Product \\ Review Generation}} 
               & ROUGE-1   & 0.182 & \textbf{0.186} \\
               & ROUGE-L   & 0.122 & \textbf{0.126} \\
               & METEOR    & 0.184 & \textbf{0.187} \\
\midrule
\multirow{3}{*}{\makecell[l]{Task $2$: Hotel \\ Experiences Generation}}
               & ROUGE-1   & 0.264 & \textbf{0.265} \\
               & ROUGE-L   & \textbf{0.152} & \textbf{0.152} \\
               & METEOR    & \textbf{0.207} & 0.206 \\
\midrule
\multirow{3}{*}{\makecell[l]{Task $3$: Stylized \\ Feedback Generation}}
               & ROUGE-1   & 0.194 & \textbf{0.205} \\
               & ROUGE-L   & 0.128 & \textbf{0.139} \\
               & METEOR    & 0.201 & \textbf{0.203} \\
\midrule
\multirow{3}{*}{\makecell[l]{Task $4$: Multilingual \\ Product Review Generation}}
               & ROUGE-1   & 0.190 & \textbf{0.191} \\
               & ROUGE-L   & 0.141 & \textbf{0.142} \\
               & METEOR    & \textbf{0.174} & 0.173 \\
\bottomrule
\end{tabular}
}
\caption{Ablation study results showing the effect of retriever choice on PGraphRAG performance. Results are reported for \emph{LLaMA-3.1-8B-Instruct} and \emph{GPT-4o-mini} on the long-text generation task (Tasks 1-4).}
\label{tab:merged_ret_long}
\end{table}

%% file: tables/merged_ret_short.tex
\begin{table}[ht]
\centering
\scriptsize 
\renewcommand{\arraystretch}{1.1} 
\resizebox{\columnwidth}{!}{ 
\begin{tabular}{llcc}
\toprule
\textbf{Short Text Generation} & \textbf{Metric} & \textbf{Contriever} & \textbf{BM25} \\
\midrule
\multicolumn{2}{l}{\textbf{\emph{\lla}}} \\
\midrule
\multirow{3}{*}{\makecell[l]{Task $5$: User Product \\ Review Title Generation}} 
               & ROUGE-1   & 0.122 & \textbf{0.125} \\
               & ROUGE-L   & 0.116 & \textbf{0.119} \\
               & METEOR    & 0.115 & \textbf{0.117} \\
\midrule
\multirow{3}{*}{\makecell[l]{Task $6$: Hotel Experience \\ Summary Generation}} 
               & ROUGE-1   & 0.117 & \textbf{0.121} \\
               & ROUGE-L   & 0.110 & \textbf{0.113} \\
               & METEOR    & 0.095 & \textbf{0.099} \\
\midrule
\multirow{3}{*}{\makecell[l]{Task $7$: Stylized Feedback \\ Title Generation}} 
               & ROUGE-1   & 0.125 & \textbf{0.132} \\
               & ROUGE-L   & 0.121 & \textbf{0.128} \\
               & METEOR    & 0.122 & \textbf{0.129} \\
\midrule
\multirow{3}{*}{\makecell[l]{Task $8$: Multi-lingual Product \\ Review Title Generation}} 
               & ROUGE-1   & 0.126 & \textbf{0.131} \\
               & ROUGE-L   & 0.118 & \textbf{0.123} \\
               & METEOR    & 0.112 & \textbf{0.118} \\
\midrule
\multicolumn{2}{l}{\textbf{\emph{GPT-4o-mini}}} \\
\midrule
\multirow{3}{*}{\makecell[l]{Task $5$: User Product \\ Review Title Generation}} 
               & ROUGE-1   & \textbf{0.113} & 0.111 \\
               & ROUGE-L   & \textbf{0.108} & 0.106 \\
               & METEOR    & \textbf{0.097} & \textbf{0.097} \\
\midrule
\multirow{3}{*}{\makecell[l]{Task $6$: Hotel Experience \\ Summary Generation}} 
               & ROUGE-1   & 0.113 & \textbf{0.118} \\
               & ROUGE-L   & 0.107 & \textbf{0.112} \\
               & METEOR    & 0.080 & \textbf{0.085} \\
\midrule
\multirow{3}{*}{\makecell[l]{Task $7$: Stylized Feedback \\ Title Generation}} 
               & ROUGE-1   & 0.108 & \textbf{0.109} \\
               & ROUGE-L   & 0.106 & \textbf{0.107} \\
               & METEOR    & 0.094 & \textbf{0.096} \\
\midrule
\multirow{3}{*}{\makecell[l]{Task $8$: Multi-lingual Product \\ Review Title Generation}} 
               & ROUGE-1   & \textbf{0.108} & \textbf{0.108} \\
               & ROUGE-L   & 0.103 & \textbf{0.104} \\
               & METEOR    & \textbf{0.082} & \textbf{0.082} \\
\bottomrule
\end{tabular}
}
\caption{Ablation study results showing the effect of retriever choice on PGraphRAG performance. Results are reported for \emph{LLaMA-3.1-8B-Instruct} and \emph{GPT-4o-mini} on the short-text generation task (Tasks 5-8).}
\label{tab:merged_ret_short}
\end{table}

%% file: tables/gpt-norank-shortext.tex
\begin{table*}[h!]
\centering
\begin{adjustbox}{width=\textwidth}
\begin{tabular}{llcccccccc}
\toprule
\textbf{Task} & \textbf{Metric} & \textbf{PGraphRAG} & \textbf{PGraphRAG*} & \textbf{PGraphRAG**} & \textbf{PGraphRAG-U} & \textbf{PGraphRAG-U*} & \textbf{PGraphRAG-U**} \\
\midrule
\multicolumn{8}{l}{\textbf{Long Text Generation}} \\
\midrule
\multirow{3}{*}{Task 1: User-Product Review Generation}
    & ROUGE-1 & 0.189 & 0.186 & \textbf{0.191} & 0.171 & 0.169 & 0.170 \\
    & ROUGE-L & \textbf{0.130} & 0.125 & \textbf{0.130} & 0.117 & 0.114 & 0.117 \\
    & METEOR  & 0.196 & 0.188 & \textbf{0.205} & 0.176 & 0.173 & 0.180 \\ 
\midrule
\multirow{3}{*}{Task 2: Hotel Experiences Generation}
    & ROUGE-1 & 0.263 & 0.266 & \textbf{0.267} & 0.221 & 0.223 & 0.225 \\
    & ROUGE-L & 0.152 & 0.152 & \textbf{0.153} & 0.135 & 0.134 & 0.135 \\
    & METEOR  & 0.206 & 0.209 & \textbf{0.216} & 0.164 & 0.168 & 0.171 \\ 
\midrule
\multirow{3}{*}{Task 3: Stylized Feedback Generation}
    & ROUGE-1 & \textbf{0.211} & 0.200 & 0.210 & 0.185 & 0.180 & 0.186 \\
    & ROUGE-L & \textbf{0.140} & 0.133 & 0.136 & 0.123 & 0.122 & 0.123 \\
    & METEOR  & 0.202 & 0.206 & \textbf{0.225} & 0.183 & 0.184 & 0.189 \\ 
\midrule
\multirow{3}{*}{Task 4: Multilingual Product Review Generation}
    & ROUGE-1 & 0.194 & 0.188 & \textbf{0.196} & 0.168 & 0.167 & 0.171 \\
    & ROUGE-L & \textbf{0.144} & 0.138 & 0.141 & 0.125 & 0.125 & 0.128 \\
    & METEOR  & 0.171 & 0.176 & \textbf{0.188} & 0.154 & 0.155 & 0.155 \\ 
\midrule
\multicolumn{8}{l}{\textbf{Short Text Generation}} \\
\midrule
\multirow{3}{*}{Task 5: User Product Review Title Generation}
    & ROUGE-1 & 0.115 & 0.114 & \textbf{0.119} & 0.108 & 0.108 & 0.111 \\
    & ROUGE-L & 0.112 & 0.109 & \textbf{0.114} & 0.105 & 0.102 & 0.105 \\
    & METEOR  & 0.099 & 0.121 & \textbf{0.128} & 0.091 & 0.116 & 0.119 \\
\midrule
\multirow{3}{*}{Task 6: Hotel Experience Summary Generation}
    & ROUGE-1 & 0.116 & 0.117 & \textbf{0.121} & 0.108 & \textbf{0.121} & 0.119 \\
    & ROUGE-L & 0.111 & 0.107 & \textbf{0.112} & 0.104 & 0.111 & 0.110 \\
    & METEOR  & 0.081 & 0.104 & \textbf{0.109} & 0.075 & \textbf{0.109} & 0.107 \\
\midrule
\multirow{3}{*}{Task 7: Stylized Feedback Title Generation}
    & ROUGE-1 & \textbf{0.122} & 0.111 & 0.120 & 0.113 & 0.115 & 0.114 \\
    & ROUGE-L & \textbf{0.118} & 0.105 & 0.114 & 0.109 & 0.109 & 0.108 \\
    & METEOR  & 0.104 & 0.117 & \textbf{0.126} & 0.096 & 0.124 & 0.123 \\
\midrule
\multirow{3}{*}{Task 8: Multi-lingual Product Review Title Generation}
    & ROUGE-1 & 0.111 & 0.108 & 0.112 & \textbf{0.115} & 0.110 & 0.110 \\
    & ROUGE-L & 0.105 & 0.100 & 0.104 & \textbf{0.107} & 0.103 & 0.101 \\
    & METEOR  & 0.083 & 0.101 & 0.105 & 0.088 & \textbf{0.108} & 0.107 \\
\bottomrule
\end{tabular}
\end{adjustbox}
\caption{Zero-shot test set results for text generation using \emph{GPT-4o-mini}. \textbf{PGraphRAG*} denotes retrieval of $k=4$ randomly selected entries without ranking, while \textbf{PGraphRAG**} represents unbounded retrieval up to the model’s context limit ($k \rightarrow \infty$).}

\label{tab:gpt-norank-text}
\end{table*}

%% file: tables/model-longtext-dev.tex
\begin{table*}[h!]
\centering
\begin{adjustbox}{width=\textwidth}
\begin{tabular}{llcccc}
\toprule
\textbf{Long Text Generation} & \textbf{Metric} & \textbf{PGraphRAG} & \textbf{LaMP} & \textbf{No-retrieval} & \textbf{Random-retrieval} \\
\midrule
\multicolumn{6}{l}{\textbf{\textit{\lla}}} \\
\midrule
\multirow{3}{*}{Task 1: User-Product Review Generation}
    & ROUGE-1 & \textbf{0.173}  & 0.168  & 0.172  & 0.126  \\
    & ROUGE-L & 0.124  & \textbf{0.125}  & 0.121  & 0.095  \\
    & METEOR  & 0.150  & 0.134  & \textbf{0.152}  & 0.101  \\ 
\midrule
\multirow{3}{*}{Task 2: Hotel Experiences Generation}
    & ROUGE-1 & \textbf{0.263}  & 0.197  & 0.224  & 0.211  \\
    & ROUGE-L & \textbf{0.156}  & 0.128  & 0.141  & 0.130  \\
    & METEOR  & \textbf{0.191}  & 0.121  & 0.148  & 0.147  \\ 
\midrule
\multirow{3}{*}{Task 3: Stylized Feedback Generation}
    & ROUGE-1 & \textbf{0.226}  & 0.181  & 0.177  & 0.142  \\
    & ROUGE-L & \textbf{0.171}  & 0.134  & 0.125  & 0.104  \\
    & METEOR  & \textbf{0.192}  & 0.147  & 0.168  & 0.119  \\ 
\midrule
\multirow{3}{*}{Task 4: Multilingual Product Review Generation}
    & ROUGE-1 & \textbf{0.174}  & 0.174  & 0.173  & 0.146  \\
    & ROUGE-L & 0.139  & \textbf{0.141}  & 0.134  & 0.117  \\
    & METEOR  & \textbf{0.133}  & 0.125  & 0.130  & 0.110  \\ 
\midrule
\multicolumn{6}{l}{\textbf{\textit{\gpt}}} \\
\midrule
\multirow{3}{*}{Task 1: User-Product Review Generation}
    & ROUGE-1 & \textbf{0.186}  & 0.169  & 0.168  & 0.157  \\
    & ROUGE-L & \textbf{0.126}  & 0.114  & 0.113  & 0.112  \\
    & METEOR  & \textbf{0.187}  & 0.170  & 0.173  & 0.148  \\ 
\midrule
\multirow{3}{*}{Task 2: Hotel Experiences Generation}
    & ROUGE-1 & \textbf{0.265}  & 0.217  & 0.222  & 0.233  \\
    & ROUGE-L & \textbf{0.152}  & 0.132  & 0.133  & 0.138  \\
    & METEOR  & \textbf{0.206}  & 0.161  & 0.164  & 0.164  \\ 
\midrule
\multirow{3}{*}{Task 3: Stylized Feedback Generation}
    & ROUGE-1 & \textbf{0.205}  & 0.178  & 0.177  & 0.168  \\
    & ROUGE-L & \textbf{0.139}  & 0.121  & 0.119  & 0.117  \\
    & METEOR  & \textbf{0.203}  & 0.178  & 0.184  & 0.160  \\ 
\midrule
\multirow{3}{*}{Task 4: Multilingual Product Review Generation}
    & ROUGE-1 & \textbf{0.191}  & 0.164  & 0.167  & 0.171  \\
    & ROUGE-L & \textbf{0.142}  & 0.123  & 0.125  & 0.131  \\
    & METEOR  & \textbf{0.173}  & 0.155  & 0.153  & 0.150  \\ 
\bottomrule
\end{tabular}
\end{adjustbox}
\caption{Zero-shot Validation set results for long text generation using \emph{\lla} and \emph{GPT-4o-mini} on Tasks 1-4.}
\label{tab:model-longtext-dev}
\end{table*}

%% file: tables/model-shorttext-dev.tex
\begin{table*}[h!]
\centering
\begin{adjustbox}{width=\textwidth}
\begin{tabular}{llcccc}
\toprule
\textbf{Short Text Generation} & \textbf{Metric} & \textbf{PGraphRAG} & \textbf{LaMP} & \textbf{No-retrieval} & \textbf{Random-retrieval} \\
\midrule
\multicolumn{6}{l}{\textbf{\textit{\lla}}} \\
\midrule
\multirow{3}{*}{Task 5: User Product Review Title Generation}
    & ROUGE-1 & \textbf{0.125}  & 0.114  & 0.111  & 0.101  \\
    & ROUGE-L & \textbf{0.119}  & 0.108  & 0.105  & 0.095  \\
    & METEOR  & \textbf{0.117}  & 0.111  & 0.104  & 0.094  \\
\midrule
\multirow{3}{*}{Task 6: Hotel Experience Summary Generation}
    & ROUGE-1 & \textbf{0.121}  & 0.119  & 0.115  & 0.115  \\
    & ROUGE-L & \textbf{0.113}  & 0.111  & 0.108  & 0.107  \\
    & METEOR  & \textbf{0.105}  & \textbf{0.105}  & 0.100  & 0.094  \\
\midrule
\multirow{3}{*}{Task 7: Stylized Feedback Title Generation}
    & ROUGE-1 & \textbf{0.132}  & 0.128  & 0.127  & 0.108  \\
    & ROUGE-L & \textbf{0.128}  & 0.124  & 0.122  & 0.104  \\
    & METEOR  & \textbf{0.129}  & 0.124  & 0.118  & 0.103  \\
\midrule
\multirow{3}{*}{Task 8: Multi-lingual Product Review Title Generation}
    & ROUGE-1 & \textbf{0.132}  & 0.128  & 0.108  & 0.127  \\
    & ROUGE-L & \textbf{0.128}  & 0.124  & 0.104  & 0.122  \\
    & METEOR  & \textbf{0.129}  & 0.124  & 0.103  & 0.118  \\
\midrule
\multicolumn{6}{l}{\textbf{\textit{\gpt}}} \\
\midrule
\multirow{3}{*}{Task 5: User Product Review Title Generation}
    & ROUGE-1 & \textbf{0.114}  & 0.106  & 0.109  & 0.107  \\
    & ROUGE-L & \textbf{0.107}  & 0.100  & 0.103  & 0.102  \\
    & METEOR  & \textbf{0.119}  & 0.115  & 0.116  & 0.109  \\
\midrule
\multirow{3}{*}{Task 6: Hotel Experience Summary Generation}
    & ROUGE-1 & \textbf{0.115}  & \textbf{0.115}  & 0.114  & 0.112  \\
    & ROUGE-L & 0.105  & \textbf{0.106}  & \textbf{0.106}  & 0.103  \\
    & METEOR  & 0.105  & \textbf{0.106}  & \textbf{0.106}  & 0.099  \\
\midrule
\multirow{3}{*}{Task 7: Stylized Feedback Title Generation}
    & ROUGE-1 & \textbf{0.105}  & 0.101  & \textbf{0.105}  & 0.098  \\
    & ROUGE-L & \textbf{0.102}  & 0.097  & 0.101  & 0.093  \\
    & METEOR  & \textbf{0.118}  & 0.111  & 0.118  & 0.105  \\
\midrule
\multirow{3}{*}{Task 8: Multi-lingual Product Review Title Generation}
    & ROUGE-1 & \textbf{0.108}  & 0.106  & \textbf{0.108}  & 0.103  \\
    & ROUGE-L & 0.099  & 0.098  & \textbf{0.099}  & 0.095  \\
    & METEOR  & 0.101  & 0.102  & \textbf{0.103}  & 0.095  \\
\bottomrule
\end{tabular}
\end{adjustbox}
\caption{Zero-shot Validation set results for short text generation using \emph{LLaMA-3.1-8B} and \emph{GPT-4o-mini} on Tasks 5-8.}
\label{tab:model-shorttext-dev}
\end{table*}

%% file: tables/model-rating-dev.tex
\begin{table*}[h!]
\centering
\begin{adjustbox}{width=\textwidth}
\begin{tabular}{llcccc}
\toprule
\textbf{Ordinal Classfication} & \textbf{Metric} & \textbf{PGraphRAG} & \textbf{LaMP} & \textbf{No-retrieval} & \textbf{Random-retrieval} \\
\midrule
\multicolumn{6}{l}{\textbf{\textit{\lla}}} \\
\midrule
\multirow{2}{*}{Task 9: User Product Review Ratings}
    & MAE \textdownarrow  & 0.3272  & 0.3220  & \textbf{0.3200}  & 0.3516  \\
    & RMSE \textdownarrow & 0.7531  & \textbf{0.7280}  & 0.7294  & 0.7972  \\
\midrule
\multirow{2}{*}{Task 10: Hotel Experience Ratings}
    & MAE \textdownarrow  & 0.3868  & 0.3685  & \textbf{0.3614}  & 0.4008  \\
    & RMSE \textdownarrow & 0.6989  & 0.6750  & \textbf{0.6643}  & 0.7178  \\
\midrule
\multirow{2}{*}{Task 11: Stylized Feedback Ratings}
    & MAE \textdownarrow  & \textbf{0.3356}  & 0.3368  & 0.3372  & 0.3812  \\
    & RMSE \textdownarrow & 0.6856  & 0.6859  & \textbf{0.6826}  & 0.7759  \\
\midrule
\multirow{2}{*}{Task 12: Multi-lingual Product Ratings}
    & MAE \textdownarrow  & 0.5228  & \textbf{0.5216}  & 0.5282  & 0.5392  \\
    & RMSE \textdownarrow & 0.8483  & \textbf{0.8395}  & 0.8519  & 0.8704  \\
\midrule
\multicolumn{6}{l}{\textbf{\textit{\gpt}}} \\
\midrule
\multirow{2}{*}{Task 9: User Product Review Ratings}
    & MAE \textdownarrow  & 0.3652  & 0.3508  & \textbf{0.3484}  & 0.4176  \\
    & RMSE \textdownarrow & 0.7125  & 0.6943  & \textbf{0.6925}  & 0.7792  \\
\midrule
\multirow{2}{*}{Task 10: Hotel Experience Ratings}
    & MAE \textdownarrow  & \textbf{0.3308}  & 0.3472  & 0.3528  & 0.3640  \\
    & RMSE \textdownarrow & \textbf{0.6056}  & 0.6394  & 0.6475  & 0.6627  \\
\midrule
\multirow{2}{*}{Task 11: Stylized Feedback Ratings}
    & MAE \textdownarrow & \textbf{0.3340}  & 0.3364  & 0.3356  & 0.3972  \\
    & RMSE \textdownarrow & 0.6515  & 0.6545  & \textbf{0.6484}  & 0.7158  \\
\midrule
\multirow{2}{*}{Task 12: Multi-lingual Product Ratings}
    & MAE \textdownarrow  & \textbf{0.4568}  & 0.4832  & 0.4908  & 0.4820  \\
    & RMSE \textdownarrow & \textbf{0.7414}  & 0.7808  & 0.7897  & 0.7917  \\
\bottomrule
\end{tabular}
\end{adjustbox}
\caption{Performance comparison on rating prediction tasks (Tasks 9-12) using \emph{GPT-4o-mini} and \emph{LLaMA-3.1-8B-Instruct} on the validation set. Results are reported using MAE and RMSE metrics across retrieval methods.}
\label{tab:model-rating-dev}
\end{table*}

%% file: text/related-work-up.tex
\section{Related Work}

\subsection*{Personalization in NLP}

Personalization in natural language processing (NLP) focuses on tailoring responses to user-specific preferences, behaviors, and contexts, improving user experience and task performance. Early work in personalized generation relied on neural encoder-decoder models and incorporated attributes such as sentiment \cite{zang-wan-2017-towards}, stylistic cues \cite{dong-etal-2017-learning-generate}, and demographic metadata \cite{huang-etal-2014-enriching}. To address data sparsity, approaches such as warm-start attention \cite{amplayo-etal-2018-cold} and user embeddings were developed.

Recent efforts have expanded personalization using retrieval-augmented generation (RAG) strategies. Methods like in-context prompting \cite{lyu2024llmrec}, retrieval-enhanced summarization \cite{richardson2023integrating}, and optimization via reinforcement learning or distillation \cite{salemi2024optimization} have improved output fluency and relevance. Benchmarking frameworks such as LaMP \cite{salemi2024lamp} and LongLaMP \cite{kumar2024longlampbenchmarkpersonalizedlongform} have standardized evaluation of personalized tasks (e.g., email writing, abstract generation). Meanwhile, retrieval-enhanced generation pipelines \cite{kim-etal-2020-retrieval} improve long-form text by incorporating relevant user history.

However, most prior work assumes dense, high-coverage user history, limiting effectiveness in cold-start or sparse-profile scenarios. Few approaches leverage structured representations (e.g., knowledge graphs) to generalize beyond individual user traces. This gap highlights a need for models that can retrieve personalized yet diverse context using structured user-item relationships.

\subsection*{Knowledge Graphs and Retrieval-Augmented Generation (RAG)}

Knowledge graphs (KGs) provide structured, relational context useful in a variety of NLP tasks such as question answering, entity linking, and reasoning \cite{liu-etal-2018-know, schneider-etal-2022-decade}. By leveraging graph traversal and multi-hop paths, KGs enable precise contextualization in tasks that require reasoning over entity relationships \cite{salnikov-etal-2023-large}. Recent techniques such as data synthesis and subgraph construction have improved KG scalability and coverage \cite{agarwal-etal-2021-knowledge}.

In parallel, retrieval-augmented generation (RAG) frameworks enhance LLMs by incorporating external memory or document retrieval into the generation process \cite{DBLP:journals/corr/abs-2007-01282}. When integrated with KGs, RAG enables structured multi-hop reasoning \cite{Saleh2024SGRAGMQ}, rare entity recognition \cite{mathur-etal-2024-doc}, and hallucination reduction in generative outputs \cite{kang2023knowledge, chen2023benchmarking}.

Despite these gains, scaling KGs in real-world systems (e.g., personalized recommendation) remains challenging \cite{Ji_2022}. Graph construction, update, and refinement require sophisticated methods to ensure correctness and completeness \cite{paulheim2017knowledge}. Moreover, traditional RAG pipelines using dense vector retrieval may struggle to integrate symbolic signals from structured graphs or handle noisy or misaligned data sources \cite{gao2024retrievalaugmented}.

\subsection*{Toward Structured Personalization via Graph-Augmented RAG}

The intersection of personalization, knowledge graphs, and RAG presents a promising research direction. Recent surveys \cite{zhang2024personalizationlargelanguagemodels} emphasize the importance of personalization in LLMs but call for approaches that generalize across users with limited history and incorporate structured context. Our work addresses this by using user-centric bipartite graphs to retrieve not only user-authored content but also related interactions from similar users, enabling robust personalization under sparse conditions.

Unlike conventional user-history-based personalization, graph-augmented RAG offers a principled way to incorporate both individual and community signals—supporting generalization, diversity, and data efficiency at inference time.